\documentclass{article}

\PassOptionsToPackage{numbers, sort&compress}{natbib}

    \usepackage[preprint]{neurips_data_2024}
\usepackage[utf8]{inputenc} % allow utf-8 input
\usepackage[T1]{fontenc}    % use 8-bit T1 fonts
\usepackage{url}            % simple URL typesetting
\usepackage{booktabs}       % professional-quality tables
\usepackage{amsfonts}       % blackboard math symbols
\usepackage{nicefrac}       % compact symbols for 1/2, etc.
\usepackage{microtype}      % microtypography

\usepackage{multirow}
\usepackage{adjustbox}
\usepackage[normalem]{ulem}
\useunder{\uline}{\ul}{}
\usepackage{graphicx}
\usepackage{amsmath}
\usepackage{tabularx}
\usepackage{xspace}
\usepackage[linesnumbered,ruled,vlined]{algorithm2e}

\usepackage{setspace}
\usepackage{multicol}
\usepackage[dvipsnames]{xcolor}
\usepackage{listings}

\usepackage{caption}

\definecolor{citecolor}{rgb}{0.21,0.49,0.74}
\definecolor{linkcolor}{rgb}{0.8,0.16,0.16}
\definecolor{codeblue}{rgb}{0.25,0.5,0.75}
\usepackage[pagebackref=true,breaklinks=true,colorlinks, citecolor=citecolor,linkcolor=linkcolor,bookmarks=false]{hyperref}

\usepackage[nottoc]{tocbibind}
\usepackage{minitoc}
\usepackage{bbm}

\makeatletter
\DeclareRobustCommand\onedot{\futurelet\@let@token\@onedot}
\def\@onedot{\ifx\@let@token.\else.\null\fi\xspace}

\def\ie{\emph{i.e}\onedot}

\makeatother

\usepackage[capitalize]{cleveref}
    \crefname{section}{Sec.}{Secs.}
    \Crefname{section}{Section}{Sections}
    \Crefname{table}{Table}{Tables}
    \crefname{table}{Table}{Tables} % I don't abbreviate tables

\makeatletter
\def\NAT@spacechar{~}% NEW
\makeatother

\usepackage{enumitem}
\setlist[itemize]{itemsep=0.1em,leftmargin=1.5em,topsep=0em}
\setlist[enumerate]{itemsep=0.2em,leftmargin=1.5em,topsep=0em}
\newcommand{\ours}{\textsc{ReXTime}\xspace}

\usepackage[nottoc]{tocbibind}
\usepackage{minitoc}
\usepackage{bbm}

\title{\ours: A Benchmark Suite for \\Reasoning-Across-Time in Videos}

\author{Jr-Jen Chen$^{1}$\quad Yu-Chien Liao$^{1}$\quad Hsi-Che Lin$^{1}$\quad Yu-Chu Yu$^{1}$\quad\\
\textbf{Yen-Chun Chen}$^{2}$\quad \textbf{Yu-Chiang Frank Wang}$^{1}$\\
$^{1}$National Taiwan University \qquad
$^{2}$Microsoft \\
{ \href{https://rextime.github.io/}{rextime.github.io}}
}

\begin{document}

% table of contents
\doparttoc % Tell to minitoc to generate a toc for the parts
\faketableofcontents % Run a fake tableofcontents command for the partocs

\maketitle

\begin{abstract}
  We introduce \ours, a benchmark designed to rigorously test AI models' ability to perform temporal reasoning within video events.
Specifically, \ours focuses on \emph{reasoning across time}, \ie human-like understanding when the question and its corresponding answer occur in different video segments.
This form of reasoning, requiring advanced understanding of cause-and-effect relationships across video segments, poses significant challenges to even the frontier multimodal large language models.
To facilitate this evaluation, we develop an automated pipeline for generating temporal reasoning question-answer pairs, significantly reducing the need for labor-intensive manual annotations.
Our benchmark includes 921 carefully vetted validation samples and 2,143 test samples, each manually curated for accuracy and relevance.
Evaluation results show that while frontier large language models outperform academic models, they still lag behind human performance by a significant 14.3\% accuracy gap.
Additionally, our pipeline creates a training dataset of 9,695 machine generated samples without manual effort, which empirical studies suggest can enhance the across-time reasoning via fine-tuning.
\end{abstract}

\section{Introduction}
\label{sec:intro}

Large Language Models~(LLMs) and Multimodal Large Language Models~(MLLMs) have nearly matched human performance in various language and vision-language tasks~\citep{achiam2023gpt,team2023gemini,Clude3}.
Notably, frontier MLLMs trained on web-scale proprietary datasets show impressive video understanding~\citep{GPT-4o}.
However, unlike LLMs which excel in text reasoning over long sequences, the cause-effect reasoning in MLLMs, especially in understanding long video events, remains under-explored.
This capability is crucial in robotics and embodied agents~\cite{ahn2022can, padalkar2023open, szot2023large}, healthcare and medicine~\cite{zhu2020question,kiciman2023causal}, and law and policy making~\cite{kiciman2023causal}.
Despite the importance, current video-language tasks like moment retrieval~\citep{gao2017tall, lei2021detecting}, highlights detection~\citep{lei2021detecting, song2015tvsum}, dense video captioning~\citep{caba2015activitynet, yang2024vidchapters}, and video question answering~\citep{xiao2021next, li2020hero} mainly address text-visual alignment, overlooking deeper temporal reasoning challenges.
\begin{figure}[t]
  \centering
  \includegraphics[width=\textwidth]{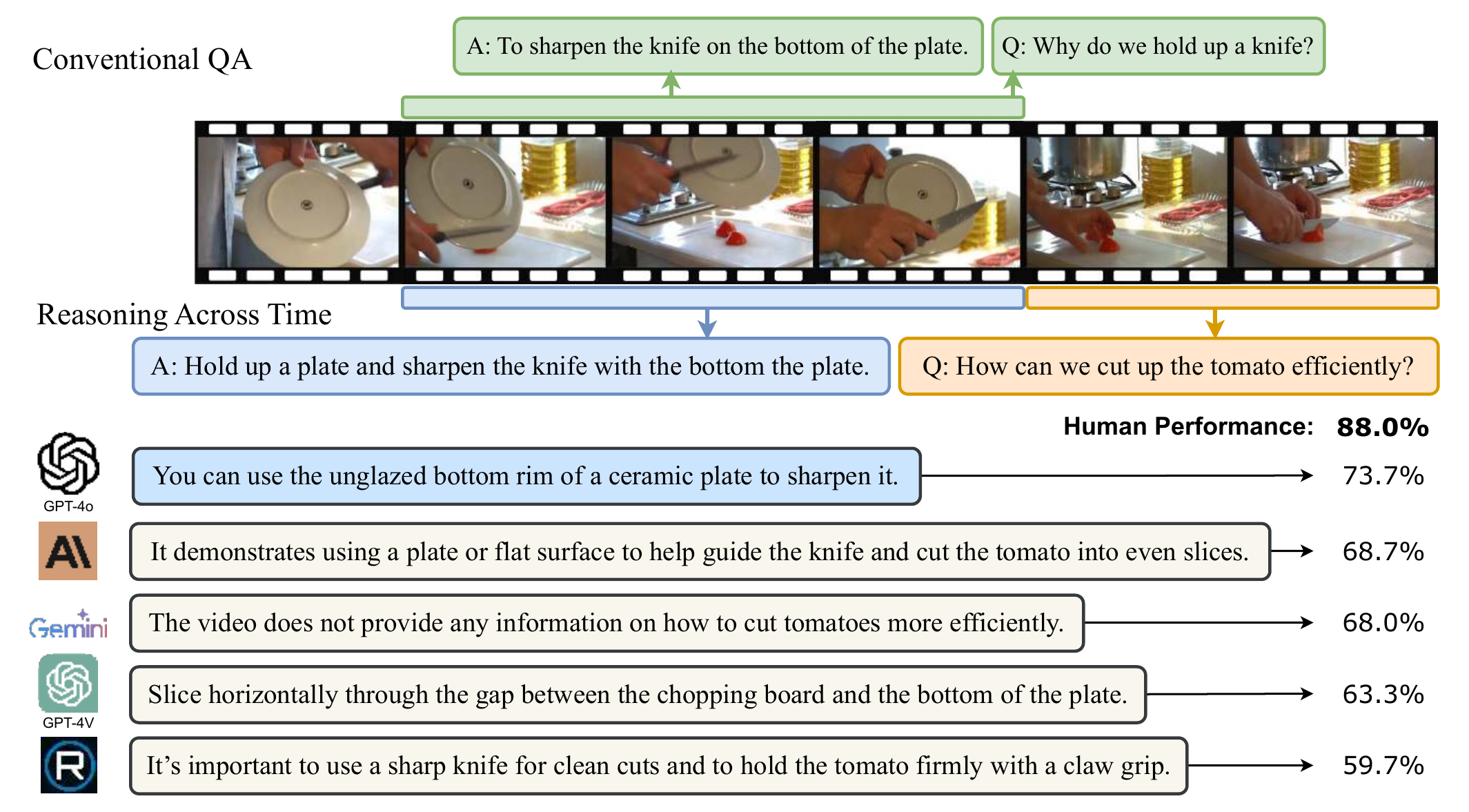}
  \caption{\textbf{A \ours example}.
  Our benchmark specializes in evaluating \emph{reasoning across time}, \ie video QA when question and answer each belong to different time spans.
  \ours poses difficulties even for frontier MLLMs, as indicated by the large gap to human-level accuracy.
  }
  \label{fig:teaser}
\end{figure}

In an initial study, we identified a common shortcoming in the most advanced MLLMs -- they struggle with video question answering when the question and answer correspond to different time segments.
As shown in~\cref{fig:teaser}, the question \textit{``How can we cut up the tomato efficiently?''} and the answer \textit{``Hold up a plate and sharpen the knife on the plate.''} each refer to separate segments.
Surprisingly, a simple question like this can challenge leading MLLMs. Therefore, there is a pressing need for a benchmark to quantitatively assess video temporal reasoning. To address this, we introduce \ours, a benchmark to evaluate \textbf{Re}asoning-A\textbf{cross}-\textbf{Time} capabilities for video events.

To develop \ours, we propose an LLM-assisted data generation pipeline that minimizes human effort and cuts costs from \$300 to \$135 per 1,000 QA pairs. The benchmark includes \textbf{921} validation and \textbf{2143} test samples, each rigorously curated by human annotators.
Empirical evidence indicates that even proprietary frontier MLLMs are inadequate for temporal reasoning.
For instance, humans can achieve 88.0\% accuracy on VQA tasks, whereas the top-performing MLLM, OpenAI's GPT-4o, only reaches 73.7\% as shown in~\cref{fig:teaser}.
A new benchmark such as \ours has the potential to significantly propel advancements in this field -- it effectively differentiates between model capabilities, and the state-of-the-art model has not yet saturated to human-level accuracy~\citep{padlewski2024vibe}.
The additional 9695 unverified samples provide a training dataset that has significantly boosted an academic MLLM's temporal reasoning skills, lowering the entry bar for future research.
Furthermore, we confirmed that \ours primarily contains \emph{reasoning across time} questions, with the lowest question-answer overlap in time~(QA-mIoU) compared to other video QA benchmarks.

To develop an efficient and effective pipeline, we have addressed two primary challenges: (1)~the quality-diversity trade-off in LLM generation, and (2)~the high cost of human labor for verification. Initially, prompting an (M)LLM to generate question-answer pairs often results in logically incorrect responses. While few-shot in-context learning enhances logical correctness, it reduces response diversity. We address this by moderating the MLLM with specific event attributes and temporal relations from a structured taxonomy. Additionally, although human verification is necessary to eliminate residual errors, we minimize costs by establishing criteria that allow the MLLM to self-assess the accuracy of its generated QAs.
As a bonus feature, we evaluate video moment localization to assess whether an AI model accurately grounds its answers to the correct video segments.

Our contributions can be summarized as the following:
\begin{itemize}
    \item \ours is the first benchmark for comprehensive video temporal reasoning focusing on cause and effect with \textbf{2143} test samples, which frontier MLLMs still lag behind human performance.
    
    \item We discover a common weakness shared by frontier MLLMs -- they reason poorly when question and answer span do not overlap.
    A newly proposed measure~\textbf{QA-IoU} quantitatively validate \ours indeed assess AI models' \emph{reasoning across time} capability.
    
    \item Our LLM-assisted data pipeline generates high quality samples with reduced human intervention, saving \textbf{55\%} of the overall cost. Furthermore, the pure machine generated training set is shown to improve the finetuning accuracy, providing a starting point for future studies.
\end{itemize}

\begin{table}[t]
\caption{\textbf{Datasets comparison}.
\ours covers features from all similar video QA tasks. Notably, \emph{reasoning-across-time} emphasizes the cause and effect understanding between visual events.
}
\label{tab:dataset_comparison}

\centering
\newcolumntype{C}{>{\centering\arraybackslash}X}
\newcolumntype{L}{>{\raggedright\arraybackslash}X}
\newcolumntype{R}{>{\raggedleft\arraybackslash}X}
\begin{tabularx}{1.0\columnwidth}{Lccccc}

\toprule
\multirow{2}{*}{Datasets} & \multirow{2}{*}{QA} & \multirow{2}{*}{Moment Localization} & \multirow{2}{*}{Training Data} & \multicolumn{2}{c}{Temporal Reasoning} \\
\cmidrule{5-6}
& & & & sequential & causal \\
\midrule
NExTQA~\citep{xiao2021next}        & $\checkmark$ &  & $\checkmark$ & $\checkmark$ &  \\
NExTGQA~\citep{xiao2023can}       & $\checkmark$ & $\checkmark$ &  & $\checkmark$ &  \\
Ego4D-NLQ~\citep{grauman2022ego4d}     &  & $\checkmark$ & $\checkmark$ & $\checkmark$ &  \\
QVHighlights~\citep{lei2021detecting}  &  & $\checkmark$ & $\checkmark$ &  &  \\
\ours          & $\checkmark$ & $\checkmark$ & $\checkmark$ & $\checkmark$ & $\checkmark$ \\
\bottomrule

\end{tabularx}
\end{table}
\section{Related work}

\paragraph{Temporal reasoning and event localization in videos}
In~\Cref{tab:dataset_comparison}, we compare \ours with related datasets on temporal reasoning or moment localization, highlighting our uniqueness. NExTQA~\citep{xiao2021next}, enhancing video understanding by explaining temporal actions, specializes in temporal reasoning but not moment localization. NExTGQA~\citep{xiao2023can}, extends NExTQA with over 10.5K temporal grounding labels, revealing models' inadequacies in grounding answers despite strong QA performance.
Ego4D-NLQ~\citep{grauman2022ego4d} lacks QA, making it difficult to assess modern AI chat assistants.
QVHighlights~\citep{lei2021detecting} featuring over 10,000 YouTube videos across various themes, aiding systems in identifying relevant moments and highlights in response to user queries. However, it does not include temporal reasoning or QA pairs.
Another related yet orthogonal work is EgoSchema~\citep{mangalam2024egoschema}, an extension of Ego4D, benchmarks long video comprehension and introduces the ``certificate length'' to measure intrinsic temporal complexity.

\paragraph{Query depend moment retrieval}
Video moment retrieval involves retrieving specific video segments based on user text queries.
Proposal-based methods~\citep{gao2017tall, anne2017localizing, hendricks2018localizing, chen2018temporally, xu2019multilevel, zhang2020learning} use a two-stage process: generate candidate proposals by scanning the entire video and then rank them based on query alignment.
In contrast, proposal-free methods~\citep{yuan2019find, zhang2020span, lin2023univtg} directly predict start and end timestamps or a center timestamp and span length.
Recent approaches integrate the Detection Transformer~(DETR)~\citep{carion2020end}, leveraging its highlight detection capabilities~\citep{lei2021detecting, moon2023query, jang2023knowing, moon2023correlation}. While these works focus on aligning visual and textual content, our research emphasizes temporal reasoning in scenarios with differing question and answer spans, requiring a distinct approach

\paragraph{Grounding large video-language models}
In the evolving landscape of Multi-modal Large Language Models~\citep{dai2024instructblip, liu2024visual, ye2023mplug, zhu2023minigpt, achiam2023gpt, team2023gemini}, significant strides have been made in the realm of video understanding~\citep{zhang2023video, maaz2023video, li2023videochat, zhang2023llama, zhu2023languagebind}, particularly in the aspect of temporal localization~\citep{huang2023vtimellm, qian2024momentor, wang2024hawkeye, ren2023timechat, huang2024lita}. VTimeLLM~\citep{huang2023vtimellm} excels with its boundary-aware training, improving Temporal Video Grounding and Dense Video Captioning. Momentor~\citep{qian2024momentor}, using the Moment-10M dataset, enhances segment-level reasoning and localization, showcasing fine-grained temporal comprehension. HawkEye~\citep{wang2024hawkeye} focuses on complex videos with time-aware objectives and innovative segment representations, achieving notable performance gain in temporal video grounding. TimeChat~\citep{ren2023timechat} uses a timestamp-aware frame encoder and flexible video token generator for better long video understanding and zero-shot temporal reasoning. LITA~\citep{huang2024lita} introduces time and SlowFast tokens~\citep{feichtenhofer2019slowfast}, significantly improving temporal localization and video-text generation. These models collectively advance temporal understanding of multimodal AI. While they claim advanced temporal reasoning, there is no quantitative evaluation. To bridge this gap, we develop a comprehensive benchmark and dataset specifically designed to evaluate and enhance the temporal reasoning ability.

\begin{figure}[t]
  \centering
  \includegraphics[width=\textwidth]{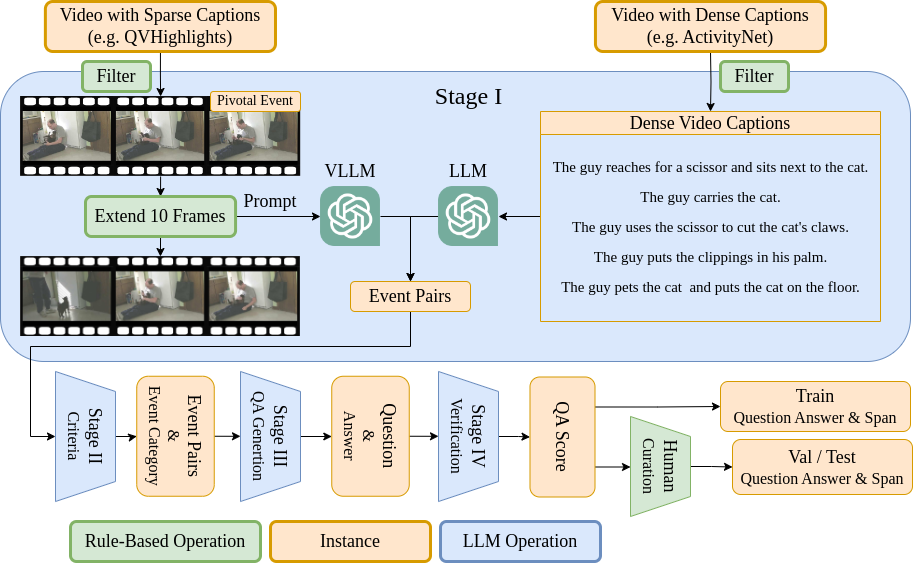}
  \caption{\textbf{Overview of the data collection pipeline.} In stage~I, we collect event pairs from two video sources. In stage~II, we score and categorize the event pairs into four relation types. In stage~III, the (M)LLM generates a question-answer pair by our carefully written few-shot demonstrations. In stage~IV, the LLM self-evaluates the generated samples to reduce the human verification cost.
  }
  \label{fig:main}
\end{figure}
\section{Data collection}

\label{sec:data_collecting}

We aim to collect video question-answer pairs to assess the \emph{reasoning-across-time} capability of multimodal AI models. A conversation involves ``reasoning-across-time'' if the question's time span does not completely overlap with the answer's time span.
By utilizing large language models and large vision language models, we create the benchmark, \ours, with much less human effort.

\subsection{Selecting videos to annotate}
\label{subsec:data_filtering}

We consider video sources with time-aligned captions (\ie, captions with start and end timestamps describing specific video segments) as they provide natural language descriptions of visual events crucial for video QA. We select ActivityNet~\citep{caba2015activitynet} and QVHighlights~\citep{lei2021detecting} datasets, which meet this criterion, for QA data creation. To ensure the QAs focus on interesting events and involve reasoning across time, we apply rule-based filtering to retain only videos that: (1)~contain at least two non-overlapping events, and (2)~have events dense enough to cover the entire video duration. Further details on the filtering process are provided in the supplementary material.

\begin{figure}[t]
  \centering
  \includegraphics[width=\textwidth]{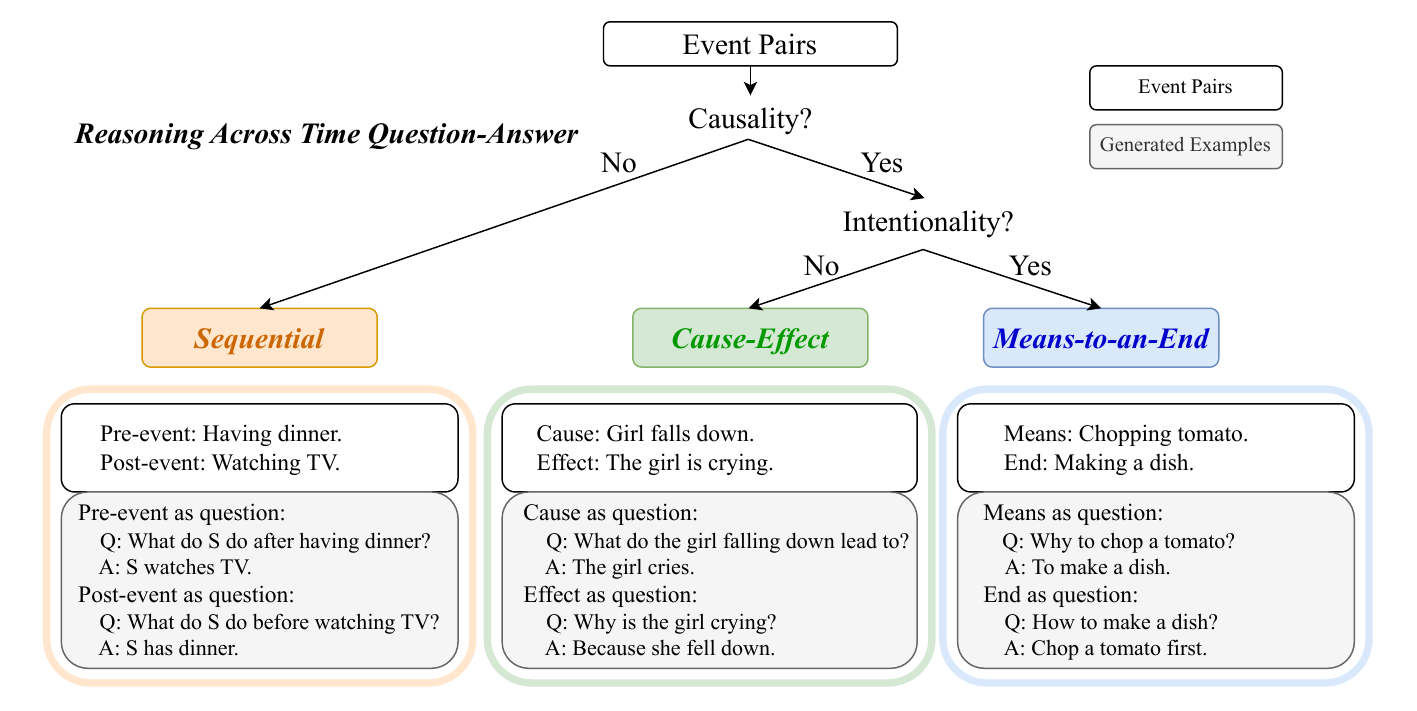}
  \caption{\textbf{Reasoning across time question-answer types} presents the relationship and examples between the three categories of question we generated. ``Having dinner / Watching TV'' does not have strong causality and is classified in \textit{sequential}, which often results in before / after questions. ``Girls falls down'' shows strong causality with ``The girl is crying.'' but lacks human intention, is classified in \textit{cause-effect}. ``Chopping tomato / Making a dish'' not only has strong causal relations but also shows subjective deliberation, which is classified into \textit{means-to-an-end}.
  }
  \label{fig:venn}
\end{figure}
\subsection{Question-answering on two events across time}

Naively feeding a video and its time-aligned captions to an MLLM often results in logically incorrect responses. Writing few-shot demonstrations improves correctness due to LLMs' strong in-context learning abilities but unexpectedly reduces diversity. To balance quality and diversity, grounding LLM generation in specific visual events and their relationships is essential. We extract event pairs from captions and categorize them into three relation types: \emph{means-to-an-end}, \emph{cause-effect}, and \emph{sequential}. Means-to-an-end refers to one event causing another with subjective intentions, \ie, ``making a dish'' leading to ``chopping tomatoes.'' Cause-effect involves causal relations without a purpose, such as ``girl falls down'' causing ``girl is crying.'' Sequential events are those with a ``before~/~after'' relation, where events do not completely overlap in time.

\paragraph{Finding candidate event pairs}
For QVHighlights videos, due to sparsely annotated captions~(events), we use MLLM to find related events given an initial ``pivotal event''. We define a caption and its annotated time span as a ``pivotal event'' and crop the corresponding video clip with $10$~second extensions before and after. This extended clip is processed by GPT-4V to detect both the cause leading to the pivotal event and its consequent effects.

For ActivityNet videos, where events~(captions) are denser, we use language-only GPT-4 to extract event pairs. We prompt the LLM to extract pairs with distinct timestamps and potential causal relations. These pairs are chosen based on their strong causal relationships, ensuring the events are temporally separated but intricately connected in terms of cause and effect.

To avoid selecting semantically identical events as candidate pairs, we ask the LLM to output a \textit{similarity score} between events and only consider less similar pairs. For detailed prompts to GPT-4V and GPT-4, please see the supplementary material.

\paragraph{Event relation classification}
We classify event pairs into the three aforementioned relations using the following four scoring criteria:
\begin{itemize}

\item \textit{Directness}: This criterion assesses the directness of the causal link between events. For example, ``A girl falls down. / She is crying.'' scores high in directness, while ``A man has dinner. / He watches TV after dinner.'' scores low.

\item \textit{Necessity}: This criterion measures whether the second event is inevitable due to the first, \ie, if the second event would still occur without the first. For example, ``The marching band aligns in the street with instruments. / A man passes in front of the marching band holding a camera.'' scores high on \textit{Directness}, but the second event is not necessarily a consequence of the first, resulting in a low \textit{Necessity} score.

\item \textit{Intentionality}: This criterion evaluates whether an event was carried out with deliberate intention. Higher scores are given when there is clear evidence of premeditated action leading to the outcome. For example, ``Chop tomato. / Making a dish.'' scores high in \textit{Intentionality} because the human intention is clear.

\item \textit{Purpose}: Even if the preceding event is executed with intention, the resulting event may not align with the original expectation. We ask the LLM to specifically detect whether the intention has been fulfilled. For example, ``Adding ingredients into a cup. / Putting a drink on the table.'' scores high in \textit{Intentionality} but low in \textit{Purpose} because the original goal was to make a drink, not to place it on a table.

\end{itemize}

We leverage GPT-4 to annotate these four scores $\in [0, 1, 2, 3]$ for each event pair. The relation can be classified using the following rules: (1)~If the sum of directness and necessity scores is below~$4$, they are in a simple \emph{sequential} relation.\footnote{We further remove the pair if the two events are not consecutive to avoid answer ambiguity, \ie, for ``before~/~after'' questions, we only consider the immediate preceding~/~following event.} (2)~If the sum of intentionality and purpose is less than~$5$, they are classified as a \emph{cause-effect} relation. (3)~If neither of the above conditions is met, the events are in a \emph{means-to-an-end} relation. \Cref{fig:venn} illustrates this process.

\paragraph{Question-answer generation}
To generate QA pairs from the LLM, we crafted \textit{in-context learning}~\citep{dong2022survey} (ICL) examples specific to each event relation (see the ICL demonstrations in the supplementary material). To create a fair benchmark that can be automatically evaluated with reliable metrics, we made \ours a multiple-choice QA task. Thus, we need to generate negative options in addition to the ground truth answer. This is easily done with a language-only LLM, and the detailed prompt is provided in the supplementary material.

\subsection{Balancing cheap machine generated data and high-quality human annotation}

\paragraph{Automatic data verification for cost reduction}
To ensure a high-quality benchmark, the correctness of the QA pairs is crucial, and a large sample size is needed to reduce variance in model evaluation. Therefore, we use LLMs to generate extensive data at a low cost, with human judges verifying the correctness of the output, which is faster than manual QA creation. To further reduce the rejection rate of LLM responses, we ask the LLM to self-verify the logical correctness of its outputs for cause-effect and means-to-an-end relationships (for sequential relations, the success rate is already high). Details of the prompts are provided in the supplementary materials. This step effectively reduces the human verification workload by filtering out poor samples. Due to the low access barrier of advanced LLMs, we generated more data than we could manually verify. Unverified data samples are used as the training dataset for \ours, serving as a jump-start dataset for future models to tackle our benchmark.

\paragraph{Mitigating the modality misalignment}
A weakness of multiple-choice QA is that AI models can learn language-only shortcuts to achieve high accuracy. To address this, we require models to output the corresponding time span of the chosen answer. A stricter metric, accuracy with IoU @ $0.5$, may better reflect true multimodal understanding ability.
One issue is that the annotated caption time spans from the original video corpus may not be accurate. Therefore, we request human annotators to re-annotate the event spans. The annotators are responsible for assessing each question-answer pair to ensure logical coherence and alignment with the video content, and for labeling the time span of the answer event.

\section{Benchmark}
\subsection{Evaluation metrics}
To evaluate performance, we use accuracy to assess multiple-choice VQA, where each question has four answer options. Additionally, we measure the model's ability to localize the answer event span using moment retrieval metrics, following~\citet{lei2021detecting}. We evaluate Recall@1 with Intersection over Union~(IoU) thresholds of $0.3$ and $0.5$ at various thresholds. A model capable of multimodal understanding should excel in both VQA and localization, with accuracy @ IoU $\geq 0.5$~\citep{xiao2023can} being a key indicator.

\begin{table}[t]
\caption{\textbf{Performances of human and frontier multi-modal large language models on the mini-test split~(300 samples).} We randomly sampled 100 examples from each event relation category and evaluated API-based frontier MLLMs. Results show that while frontier MLLMs show certain degrees of temporal reasoning, they struggle with moment localization. We also estimate human-level performance, where each question is answered by three workers. The finding reveals that recent MLLMs are still far behind humans in both temporal reasoning VQA and moment localization.
}

\centering
\newcolumntype{C}{>{\centering\arraybackslash}X}
\newcolumntype{L}{>{\raggedright\arraybackslash}X}
\newcolumntype{R}{>{\raggedleft\arraybackslash}X}
\begin{tabularx}{\columnwidth}{lCCCCC}
\toprule
\multirow{3}{*}{Models} & \multicolumn{3}{c}{Moment Localization} &  \multicolumn{2}{c}{VQA} \\
\cmidrule(lr){2-4}
\cmidrule(lr){5-6}
       & \multirow{2}{*}{mIoU} & R@1  & R@1  & Accuracy(\%) & Accuracy(\%)\\
       & & \footnotesize (IoU=0.3) & \footnotesize (IoU=0.5) &  & \footnotesize @IoU $\geq 0.5$\\
\midrule
Human & \textbf{61.11} & \textbf{74.30} & \textbf{62.85} & \textbf{87.98} & \textbf{58.51} \\
\cmidrule(lr){1-6}
GPT-4o~\citep{GPT-4o} & \textbf{36.28} & \textbf{45.33} & \textbf{34.00} & \textbf{73.67} & \textbf{28.67} \\
Claude3-Opus~\citep{Clude3} & 23.61 & 30.67 & 17.67 & 68.67 & 13.67\\
Gemini-1.5-Pro~\citep{team2023gemini} & 28.43 & 35.67 & 25.00 & 68.00 & 18.33 \\
GPT-4V~\citep{achiam2023gpt} & 26.74 & 33.33 & 22.00 & 63.33 & 16.67 \\
Reka-Core~\citep{Reka} & 27.95 & 36.33 & 24.00 & 59.67 & 17.00 \\
\bottomrule
\end{tabularx}
\label{tab:human_performance}
\end{table}
\subsection{How far are frontier MLLMs to solving \ours?}
\label{subsec:human_eval}
\Cref{tab:human_performance} shows the performance of humans and various multi-modal large language models, including GPT-4V~\citep{achiam2023gpt}, GPT-4o~\citep{GPT-4o}, Gemini~\citep{team2023gemini}, Claude~\citep{Clude3}, and Reka~\citep{Reka}. For evaluating MLLMs, we prompt the models to predict the time span directly and select the most likely options. Detailed settings for each model are provided in the supplementary materials. Due to budget constraints and API query limits, we used a mini-test split of $300$ samples. Human-level performance is included to set a benchmark for AI models and to identify future benchmark saturation.

In conclusion, the leading VLLMs can reason across time to some extent, as shown in the VQA accuracy. The newest MLLM, Reka, achieves $59.67\%$, and the best model, GPT-4o, achieves $73.67\%$. However, these models still lag behind the human-level accuracy of~$87.98\%$. Despite claims of strong vision capabilities, these models often fail to localize the correct answer span, resulting in significantly lower mIoU compared to human performance.

\subsection{Are academic and open source models competitive?}
We consider both moment localization models~\citep{lin2023univtg, moon2023correlation} and LLM-based models~\citep{huang2023vtimellm, ren2023timechat, huang2024lita}, and evaluate both zero-shot~(\cref{tab:zero-shot}) and fine-tuned performance~(\cref{tab:finetuned}).
A key observation is that most current open-source models struggle to accurately localize the ground truth moment in \ours. Compared to proprietary frontier models, the zero-shot VQA accuracy of these open-source models is significantly lower. For pure VQA on temporal reasoning, humans can achieve $87.98\%$ accuracy, the best proprietary API achieves $73.67\%$, and the best open-source model only achieves $38.45\%$ accuracy.
As contrasted, models trained on our dataset, as shown in~\cref{tab:finetuned}, perform better on the moment retrieval task compared to the best proprietary API. The best-performing model, UniVTG, achieves an mIoU of $34.73\%$, which is competitive with frontier models at $36.28\%$. This indicates that frontier MLLMs are still not well-equipped for moment retrieval. Last but not least, we can see that after trained on our dataset, VTimeLLM gets a significant improvement from $36.25\%$ to $58.15\%$ on VQA. This result is even comparable to a frontier MLLM -- Reka. Similarly, TimeChat improves from $38.45\%$ to $49.35\%$.
Moreover, open source grounding language models can get a significant improvement on moment localization.
In conclusion, utilizing our automatic generation pipeline, we can generate training data both effectively and efficiently with less than $10\%$ of the manual annotation cost in~(see supplementary for detailed calculations).
This could serve as a good starting point for future multimodal models' improvement on temporal reasoning.

\begin{table}[t]
\caption{\textbf{Zero-shot performance of open source models on the test split.} We assess the zero-shot capabilities of state-of-the-art moment retrieval models and grounding video LLMs. We choose two non-generative vision-language models~\citep{lin2023univtg, moon2023correlation} and three LLM-based methods~\citep{huang2023vtimellm, ren2023timechat, huang2024lita} with publicly available code and model weights.
We can see open source models significantly lag behind frontier LLMs in temporal reasoning VQA.
}
\centering
\newcolumntype{C}{>{\centering\arraybackslash}X}
\newcolumntype{L}{>{\raggedright\arraybackslash}X}
\newcolumntype{R}{>{\raggedleft\arraybackslash}X}
\begin{tabularx}{\columnwidth}{lRRRC}
\toprule
\multirow{3}{*}{Models} & \multicolumn{3}{c}{Moment Localization} &  \multirow{3}{*}{VQA} \\
\cmidrule(lr){2-4}
       &  \multirow{2}{*}{mIoU} & R@1 $\,\:$ & R@1 $\,\:$ &  \\
       & & \footnotesize (IoU=0.3) & \footnotesize (IoU=0.5) &  \\
\midrule
UniVTG~\citep{lin2023univtg}   &  \textbf{28.17} & \textbf{41.34} $\,\:$   & \textbf{26.88}  $\,\:$& $-\;\,$ \\
CG-DETR~\citep{moon2023correlation}    & 23.87 & 31.31 $\,\:$ & 16.67 $\,\:$ & $-\;\,$ \\
\cmidrule(lr){1-5}
VTimeLLM~\citep{huang2023vtimellm}    & 20.14 & 28.84 $\,\:$ & 17.41  $\,\:$& 36.16 \\
TimeChat~\citep{ren2023timechat}    & 11.65 & 14.42 $\,\:$ & 7.61  $\,\:$ & \textbf{40.04} \\
LITA~\citep{huang2024lita}       & 21.49 & 29.49  $\,\:$ & 16.29 $\,\:$ & 34.44 \\
\bottomrule
\end{tabularx}
\label{tab:zero-shot}
\end{table}

\begin{table}[t]
\caption{\textbf{Test set performance of open source models after finetuning.}
The results show that our fully automatic pipeline may provide useful training data to tech models to reason across time.
We skip LITA~\citep{huang2024lita} because the only publicly accessible model contains 13B parameters, which is beyond our computation resource to finetune. 
}
\centering
\newcolumntype{C}{>{\centering\arraybackslash}X}
\newcolumntype{L}{>{\raggedright\arraybackslash}X}
\newcolumntype{R}{>{\raggedleft\arraybackslash}X}
\begin{tabularx}{\columnwidth}{lCCCcc}
\toprule
\multirow{3}{*}{Models} & \multicolumn{3}{c}{Moment Localization} &  \multicolumn{2}{c}{VQA} \\
\cmidrule(lr){2-4}
\cmidrule(lr){5-6}
       & \multirow{2}{*}{mIoU} & R@1  & R@1  & Accuracy(\%) & Accuracy(\%)\\
       && \footnotesize (IoU=0.3) & \footnotesize (IoU=0.5) &  & \footnotesize @ IoU $\geq 0.5$\\
\midrule
UniVTG~\citep{lin2023univtg}   & \textbf{34.63} & \textbf{53.48} & \textbf{34.53} & $-$ & $-$\\
CG-DETR~\citep{moon2023correlation}  & 26.53 & 39.71 & 22.73 & $-$ & $-$\\
\cmidrule(lr){1-6}
VTimeLLM~\citep{huang2023vtimellm} & 29.92 & 43.69 & 26.13 & \textbf{57.58} & \textbf{17.13}\\
TimeChat~\citep{ren2023timechat}  & 26.29 & 40.13 & 21.42 & 49.46 & 10.92\\

% LITA~\citep{huang2024lita}     &  &  &  &  &  & \\
\bottomrule
\end{tabularx}
\label{tab:finetuned}
\end{table}

\begin{table}[t]  % Table floats to the top of the page
\caption{\textbf{Dataset statistics}. Our comparison focuses on datasets with both question queries and moment localization features.
We present a comprehensive report detailing the number of temporal reasoning samples on each split, certificate length~(C.L.) and Question-Answer mean Intersection over Union~(QA-mIoU) respectively. A higher average certificate length indicates that a model needs to reason across a longer duration in a video. A lower QA-mIoU indicates smaller intersection of question span and answer span, requiring the model to reason across different time segments in a video.
From the qualitative measures, \ours serves as a better benchmark to evaluate the reasoning across time capability.
($^\dagger$: Only counts temporal reasoning QA pairs. See supplementary for details.)
}
\label{tab:dataset_statistics}
\centering

\newcolumntype{C}{>{\centering\arraybackslash}X}  % Centered column
\newcolumntype{L}{>{\raggedright\arraybackslash}X}  % Left aligned column
\newcolumntype{R}{>{\raggedleft\arraybackslash}X}  % Right aligned column

\begin{tabularx}{\columnwidth}{lRRRRc}  % Define the column alignments and widths

\toprule
\multirow{2}{*}{Datasets} & \multicolumn{3}{c}{\# of Reasoning Across Time Samples} & \multirow{2}{*}{C.L. (s) $\uparrow$} & \multirow{2}{*}{QA-mIoU (\%) $\downarrow$} \\
\cmidrule{2-4}
 & Train$\;\;$ & Val$\;\;$ & Test$\;\;$ & & \\
\midrule
Ego4D-NLQ~\citep{grauman2022ego4d}  & 2,212$^\dagger$ & 775$^\dagger$ & 705$^\dagger$ & 5.2 $\quad$ & 85.5 \\
NExTGQA~\citep{xiao2023can}  & $-\quad\;$ & 1,403$^\dagger$ & 2,301$^\dagger$ & 11.7 $\quad$ & 66.1 \\
\ours  & 9,695$\;\,$ & 921$\;\,$ & 2,143$\;\,$ & \textbf{66.0} $\quad$ & \textbf{15.5} \\
\bottomrule
\end{tabularx}

\end{table}
\begin{figure}[t]
  \centering
  \includegraphics[width=\textwidth]{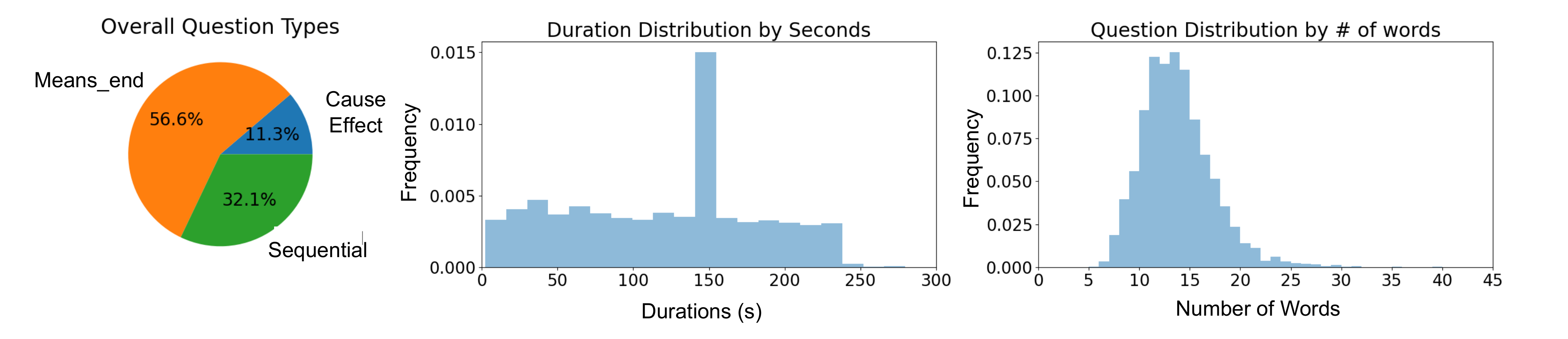}
  \caption{\textbf{Data distribution.} We visualize the distribution of our collected question-answer pairs.
  The pie chart shows the overall percentage of each relation category.
  The middle histogram shows the distribution of the number of words in a question.
  The right histogram shows the video duration distribution.
  The lower number of \textit{Cause-Effect} samples in ActivityNet can be attributed to the nature of the dataset, which predominantly features human activities.
  These activities typically involve deliberate actions with specific intentions, leading to a higher percentage of \textit{Means-to-an-End} instances.
  }
  \label{fig:charts}
\end{figure}
\subsection{Dataset statistics}

\paragraph{Question-answer intersection of union}
To quantify ``across-time'' reasoning, we introduce a new measure called Question-Answer Intersection over Union~(QA-IoU). QA-IoU is calculated by dividing the intersection of the time spans of the question and answer by their union. A lower QA-IoU indicates a greater need for reasoning across time, as it reflects smaller time overlaps between the question and answer spans. To excel in a low QA-m(ean)IoU video question-answering task, a model must understand the temporal relationships between events, presenting significant challenges to modern multimodal AI assistants.

\paragraph{Average certificate lengths}
\citet{mangalam2024egoschema} defined Certificate Length (C.L.) as the minimal length of the video segment necessary to answer a given question. In \ours, C.L. corresponds to the interval from the earliest start timestamp to the latest end timestamp of the question and answer spans. A longer Certificate Length requires the model to consider a longer segment to answer the question, increasing the difficulty for AI models.

\paragraph{Comparison to similar tasks}
\label{benchmark:dataset_comparison}
Ego4D-NLQ is a task under the Ego4D Challenge~\citep{grauman2022ego4d} in the Episodic Memory category.\footnote{\url{https://ego4d-data.org/docs/challenge/}.}
Given a video clip and a natural language query, Ego4D-NLQ requires a model to localize the temporal window within the entire video history where the answer to the question is evident. NExTGQA~\citep{xiao2023can} extends NExT-QA~\citep{xiao2021next} with $10.5$k temporal grounding (or location) labels tied to the original QA pairs.

We compare \ours to the above two datasets on the number of reasoning across time samples, certificate length, and QA-mIoU. As depicted in \cref{tab:dataset_statistics}, the average certificate length in our dataset is considerably longer than in existing tasks. This suggests that effectively addressing our task requires models to have more advanced temporal reasoning abilities.

The lower QA-mIoU in \ours indicates that an AI model needs to first locate the question event and then scan the rest of the visual events in the video to reason about the correct answer. This is more challenging because the reasoning and moment localization cannot be easily decomposed. For existing tasks, a model mostly needs to localize the question event and then reason within roughly the same span due to the higher QA-IoU.

Note that EgoSchema~\citep{mangalam2024egoschema}, which also poses significant challenges to modern deep learning systems, would be measured the longest certificate length mainly because its questions often ask for average statistics or total counts of event occurrences throughout the video. Since this is not related to our focus on long-distance event relational reasoning, we do not include it in the table.

\paragraph{Other statistics}
\Cref{fig:charts} provides additional analysis on question types, the distribution of question lengths in words, and video durations. We emphasize that \ours is diverse, as simple ``before/after'' questions account for less than $40\%$ of the dataset, and a significant portion of the questions contain more than $15$ words. Additionally, most videos are longer than $100$ seconds, posing a challenging test for the multimodal model's ability to handle long sequences.

\section{Conclusion}
We propose \ours, a comprehensive and reliable benchmark for multimodal AI, emphasizing \emph{reasoning-across-time} and visual event localization in videos, with minimal human labor. We demonstrate that even frontier MLLMs found \ours difficult and fall far behind human-level performance. The automatically constructed training dataset further points out a promising way for future models to equip the capability.

\section*{Acknowledgement}
We express our profound gratitude to all volunteers who contributed to the human annotation process for this project. The efforts have been essential in ensuring the data's accuracy and reliability, and our project's success could not have been realized without the commitment.
Special thanks to Sheng-Yu Huang, I-Jieh Liu, Zi-Ting Chou, Bin Shih Wu, Yu Chi Pai, Yu-Chen Cho, Wei Wang, Yu Hsiang Huang, Hsin Chen Lin, Hung Kai Chung, Kuei Chun Wang, and Ko An Chu for your support and collaboration pivotal in achieving our project's goals.

\renewcommand{\bibname}{References}
{
\small
\bibliography{references} 

\begin{thebibliography}{50}
\providecommand{\natexlab}[1]{#1}
\providecommand{\url}[1]{\texttt{#1}}
\expandafter\ifx\csname urlstyle\endcsname\relax
  \providecommand{\doi}[1]{doi: #1}\else
  \providecommand{\doi}{doi: \begingroup \urlstyle{rm}\Url}\fi

\bibitem[Clu(2024)]{Clude3}
The claude 3 model family: Opus, sonnet, haiku.
\newblock Technical report, Anthropic, 2024.

\bibitem[GPT(2024)]{GPT-4o}
Gpt-4 system card.
\newblock Technical report, OpenAI, 2024.

\bibitem[Rek(2024)]{Reka}
Reka core, flash, and edge: A series of powerful multimodal language models.
\newblock Technical report, Reka, 2024.

\bibitem[Achiam et~al.(2023)Achiam, Adler, Agarwal, Ahmad, Akkaya, Aleman, Almeida, Altenschmidt, Altman, Anadkat, et~al.]{achiam2023gpt}
Josh Achiam, Steven Adler, Sandhini Agarwal, Lama Ahmad, Ilge Akkaya, Florencia~Leoni Aleman, Diogo Almeida, Janko Altenschmidt, Sam Altman, Shyamal Anadkat, et~al.
\newblock Gpt-4 technical report.
\newblock \emph{arXiv preprint arXiv:2303.08774}, 2023.

\bibitem[Ahn et~al.(2022)Ahn, Brohan, Brown, Chebotar, Cortes, David, Finn, Fu, Gopalakrishnan, Hausman, et~al.]{ahn2022can}
Michael Ahn, Anthony Brohan, Noah Brown, Yevgen Chebotar, Omar Cortes, Byron David, Chelsea Finn, Chuyuan Fu, Keerthana Gopalakrishnan, Karol Hausman, et~al.
\newblock Do as i can, not as i say: Grounding language in robotic affordances.
\newblock \emph{arXiv preprint arXiv:2204.01691}, 2022.

\bibitem[Anne~Hendricks et~al.(2017)Anne~Hendricks, Wang, Shechtman, Sivic, Darrell, and Russell]{anne2017localizing}
Lisa Anne~Hendricks, Oliver Wang, Eli Shechtman, Josef Sivic, Trevor Darrell, and Bryan Russell.
\newblock Localizing moments in video with natural language.
\newblock In \emph{ICCV}, 2017.

\bibitem[Caba~Heilbron et~al.(2015)Caba~Heilbron, Escorcia, Ghanem, and Carlos~Niebles]{caba2015activitynet}
Fabian Caba~Heilbron, Victor Escorcia, Bernard Ghanem, and Juan Carlos~Niebles.
\newblock Activitynet: A large-scale video benchmark for human activity understanding.
\newblock In \emph{CVPR}, 2015.

\bibitem[Carion et~al.(2020)Carion, Massa, Synnaeve, Usunier, Kirillov, and Zagoruyko]{carion2020end}
Nicolas Carion, Francisco Massa, Gabriel Synnaeve, Nicolas Usunier, Alexander Kirillov, and Sergey Zagoruyko.
\newblock End-to-end object detection with transformers.
\newblock In \emph{ECCV}, 2020.

\bibitem[Chen et~al.(2018)Chen, Chen, Ma, Jie, and Chua]{chen2018temporally}
Jingyuan Chen, Xinpeng Chen, Lin Ma, Zequn Jie, and Tat-Seng Chua.
\newblock Temporally grounding natural sentence in video.
\newblock In \emph{EMNLP}, 2018.

\bibitem[Dai et~al.(2024)Dai, Li, Li, Tiong, Zhao, Wang, Li, Fung, and Hoi]{dai2024instructblip}
Wenliang Dai, Junnan Li, Dongxu Li, Anthony Meng~Huat Tiong, Junqi Zhao, Weisheng Wang, Boyang Li, Pascale~N Fung, and Steven Hoi.
\newblock Instructblip: Towards general-purpose vision-language models with instruction tuning.
\newblock In \emph{NeurIPS}, 2024.

\bibitem[Dong et~al.(2022)Dong, Li, Dai, Zheng, Wu, Chang, Sun, Xu, and Sui]{dong2022survey}
Qingxiu Dong, Lei Li, Damai Dai, Ce~Zheng, Zhiyong Wu, Baobao Chang, Xu~Sun, Jingjing Xu, and Zhifang Sui.
\newblock A survey on in-context learning.
\newblock \emph{arXiv preprint arXiv:2301.00234}, 2022.

\bibitem[Feichtenhofer et~al.(2019)Feichtenhofer, Fan, Malik, and He]{feichtenhofer2019slowfast}
Christoph Feichtenhofer, Haoqi Fan, Jitendra Malik, and Kaiming He.
\newblock Slowfast networks for video recognition.
\newblock In \emph{ICCV}, 2019.

\bibitem[Gao et~al.(2017)Gao, Sun, Yang, and Nevatia]{gao2017tall}
Jiyang Gao, Chen Sun, Zhenheng Yang, and Ram Nevatia.
\newblock Tall: Temporal activity localization via language query.
\newblock In \emph{ICCV}, 2017.

\bibitem[Grauman et~al.(2022)Grauman, Westbury, Byrne, Chavis, Furnari, Girdhar, Hamburger, Jiang, Liu, Liu, et~al.]{grauman2022ego4d}
Kristen Grauman, Andrew Westbury, Eugene Byrne, Zachary Chavis, Antonino Furnari, Rohit Girdhar, Jackson Hamburger, Hao Jiang, Miao Liu, Xingyu Liu, et~al.
\newblock Ego4d: Around the world in 3,000 hours of egocentric video.
\newblock In \emph{CVPR}, 2022.

\bibitem[Hendricks et~al.(2018)Hendricks, Wang, Shechtman, Sivic, Darrell, and Russell]{hendricks2018localizing}
Lisa~Anne Hendricks, Oliver Wang, Eli Shechtman, Josef Sivic, Trevor Darrell, and Bryan Russell.
\newblock Localizing moments in video with temporal language.
\newblock In \emph{EMNLP}, 2018.

\bibitem[Hu et~al.(2022)Hu, Shen, Wallis, Allen-Zhu, Li, Wang, Wang, and Chen]{hu2022lora}
Edward~J Hu, Yelong Shen, Phillip Wallis, Zeyuan Allen-Zhu, Yuanzhi Li, Shean Wang, Lu~Wang, and Weizhu Chen.
\newblock Lo{RA}: Low-rank adaptation of large language models.
\newblock In \emph{ICLR}, 2022.

\bibitem[Huang et~al.(2024{\natexlab{a}})Huang, Wang, Chen, Song, and Zhu]{huang2023vtimellm}
Bin Huang, Xin Wang, Hong Chen, Zihan Song, and Wenwu Zhu.
\newblock Vtimellm: Empower llm to grasp video moments.
\newblock In \emph{CVPR}, 2024{\natexlab{a}}.

\bibitem[Huang et~al.(2024{\natexlab{b}})Huang, Liao, Radhakrishnan, Yin, Molchanov, Yu, and Kautz]{huang2024lita}
De-An Huang, Shijia Liao, Subhashree Radhakrishnan, Hongxu Yin, Pavlo Molchanov, Zhiding Yu, and Jan Kautz.
\newblock Lita: Language instructed temporal-localization assistant.
\newblock \emph{arXiv preprint arXiv:2403.19046}, 2024{\natexlab{b}}.

\bibitem[Jang et~al.(2023)Jang, Park, Kim, Kwon, and Sohn]{jang2023knowing}
Jinhyun Jang, Jungin Park, Jin Kim, Hyeongjun Kwon, and Kwanghoon Sohn.
\newblock Knowing where to focus: Event-aware transformer for video grounding.
\newblock In \emph{ICCV}, 2023.

\bibitem[K{\i}c{\i}man et~al.(2023)K{\i}c{\i}man, Ness, Sharma, and Tan]{kiciman2023causal}
Emre K{\i}c{\i}man, Robert Ness, Amit Sharma, and Chenhao Tan.
\newblock Causal reasoning and large language models: Opening a new frontier for causality.
\newblock \emph{arXiv preprint arXiv:2305.00050}, 2023.

\bibitem[Lei et~al.(2021)Lei, Berg, and Bansal]{lei2021detecting}
Jie Lei, Tamara~L Berg, and Mohit Bansal.
\newblock Detecting moments and highlights in videos via natural language queries.
\newblock In \emph{NeurIPS}, 2021.

\bibitem[Li et~al.(2023)Li, He, Wang, Li, Wang, Luo, Wang, Wang, and Qiao]{li2023videochat}
KunChang Li, Yinan He, Yi~Wang, Yizhuo Li, Wenhai Wang, Ping Luo, Yali Wang, Limin Wang, and Yu~Qiao.
\newblock Videochat: Chat-centric video understanding.
\newblock \emph{arXiv preprint arXiv:2305.06355}, 2023.

\bibitem[Li et~al.(2020)Li, Chen, Cheng, Gan, Yu, and Liu]{li2020hero}
Linjie Li, Yen-Chun Chen, Yu~Cheng, Zhe Gan, Licheng Yu, and Jingjing Liu.
\newblock Hero: Hierarchical encoder for video+ language omni-representation pre-training.
\newblock In \emph{EMNLP}, 2020.

\bibitem[Lin et~al.(2023)Lin, Zhang, Chen, Pramanick, Gao, Wang, Yan, and Shou]{lin2023univtg}
Kevin~Qinghong Lin, Pengchuan Zhang, Joya Chen, Shraman Pramanick, Difei Gao, Alex~Jinpeng Wang, Rui Yan, and Mike~Zheng Shou.
\newblock Univtg: Towards unified video-language temporal grounding.
\newblock In \emph{ICCV}, 2023.

\bibitem[Liu et~al.(2024)Liu, Li, Wu, and Lee]{liu2024visual}
Haotian Liu, Chunyuan Li, Qingyang Wu, and Yong~Jae Lee.
\newblock Visual instruction tuning.
\newblock In \emph{NeurIPS}, 2024.

\bibitem[Maaz et~al.(2023)Maaz, Rasheed, Khan, and Khan]{maaz2023video}
Muhammad Maaz, Hanoona Rasheed, Salman Khan, and Fahad~Shahbaz Khan.
\newblock Video-chatgpt: Towards detailed video understanding via large vision and language models.
\newblock \emph{arXiv preprint arXiv:2306.05424}, 2023.

\bibitem[Mangalam et~al.(2024)Mangalam, Akshulakov, and Malik]{mangalam2024egoschema}
Karttikeya Mangalam, Raiymbek Akshulakov, and Jitendra Malik.
\newblock Egoschema: A diagnostic benchmark for very long-form video language understanding.
\newblock In \emph{NeurIPS}, 2024.

\bibitem[Moon et~al.(2023{\natexlab{a}})Moon, Hyun, Lee, and Heo]{moon2023correlation}
WonJun Moon, Sangeek Hyun, SuBeen Lee, and Jae-Pil Heo.
\newblock Correlation-guided query-dependency calibration in video representation learning for temporal grounding.
\newblock \emph{arXiv preprint arXiv:2311.08835}, 2023{\natexlab{a}}.

\bibitem[Moon et~al.(2023{\natexlab{b}})Moon, Hyun, Park, Park, and Heo]{moon2023query}
WonJun Moon, Sangeek Hyun, SangUk Park, Dongchan Park, and Jae-Pil Heo.
\newblock Query-dependent video representation for moment retrieval and highlight detection.
\newblock In \emph{CVPR}, 2023{\natexlab{b}}.

\bibitem[Padalkar et~al.(2023)Padalkar, Pooley, Jain, Bewley, Herzog, Irpan, Khazatsky, Rai, Singh, Brohan, et~al.]{padalkar2023open}
Abhishek Padalkar, Acorn Pooley, Ajinkya Jain, Alex Bewley, Alex Herzog, Alex Irpan, Alexander Khazatsky, Anant Rai, Anikait Singh, Anthony Brohan, et~al.
\newblock Open x-embodiment: Robotic learning datasets and rt-x models.
\newblock \emph{arXiv preprint arXiv:2310.08864}, 2023.

\bibitem[Padlewski et~al.(2024)Padlewski, Bain, Henderson, Zhu, Relan, Pham, Ong, Aleksiev, Ormazabal, Phua, et~al.]{padlewski2024vibe}
Piotr Padlewski, Max Bain, Matthew Henderson, Zhongkai Zhu, Nishant Relan, Hai Pham, Donovan Ong, Kaloyan Aleksiev, Aitor Ormazabal, Samuel Phua, et~al.
\newblock Vibe-eval: A hard evaluation suite for measuring progress of multimodal language models.
\newblock \emph{arXiv preprint arXiv:2405.02287}, 2024.

\bibitem[Qian et~al.(2024)Qian, Li, Wu, Ye, Fei, Chua, Zhuang, and Tang]{qian2024momentor}
Long Qian, Juncheng Li, Yu~Wu, Yaobo Ye, Hao Fei, Tat-Seng Chua, Yueting Zhuang, and Siliang Tang.
\newblock Momentor: Advancing video large language model with fine-grained temporal reasoning.
\newblock In \emph{ICML}, 2024.

\bibitem[Ren et~al.(2024)Ren, Yao, Li, Sun, and Hou]{ren2023timechat}
Shuhuai Ren, Linli Yao, Shicheng Li, Xu~Sun, and Lu~Hou.
\newblock Timechat: A time-sensitive multimodal large language model for long video understanding.
\newblock In \emph{CVPR}, 2024.

\bibitem[Song et~al.(2015)Song, Vallmitjana, Stent, and Jaimes]{song2015tvsum}
Yale Song, Jordi Vallmitjana, Amanda Stent, and Alejandro Jaimes.
\newblock Tvsum: Summarizing web videos using titles.
\newblock In \emph{CVPR}, 2015.

\bibitem[Szot et~al.(2024)Szot, Schwarzer, Agrawal, Mazoure, Metcalf, Talbott, Mackraz, Hjelm, and Toshev]{szot2023large}
Andrew Szot, Max Schwarzer, Harsh Agrawal, Bogdan Mazoure, Rin Metcalf, Walter Talbott, Natalie Mackraz, R~Devon Hjelm, and Alexander~T Toshev.
\newblock Large language models as generalizable policies for embodied tasks.
\newblock In \emph{CoRL}, 2024.

\bibitem[Team et~al.(2023)Team, Anil, Borgeaud, Wu, Alayrac, Yu, Soricut, Schalkwyk, Dai, Hauth, et~al.]{team2023gemini}
Gemini Team, Rohan Anil, Sebastian Borgeaud, Yonghui Wu, Jean-Baptiste Alayrac, Jiahui Yu, Radu Soricut, Johan Schalkwyk, Andrew~M Dai, Anja Hauth, et~al.
\newblock Gemini: a family of highly capable multimodal models.
\newblock \emph{arXiv preprint arXiv:2312.11805}, 2023.

\bibitem[Wang et~al.(2024)Wang, Meng, Liang, Wang, Liu, and Zhao]{wang2024hawkeye}
Yueqian Wang, Xiaojun Meng, Jianxin Liang, Yuxuan Wang, Qun Liu, and Dongyan Zhao.
\newblock Hawkeye: Training video-text llms for grounding text in videos.
\newblock \emph{arXiv preprint arXiv:2403.10228}, 2024.

\bibitem[Xiao et~al.(2021)Xiao, Shang, Yao, and Chua]{xiao2021next}
Junbin Xiao, Xindi Shang, Angela Yao, and Tat-Seng Chua.
\newblock Next-qa: Next phase of question-answering to explaining temporal actions.
\newblock In \emph{CVPR}, 2021.

\bibitem[Xiao et~al.(2024)Xiao, Yao, Li, and Chua]{xiao2023can}
Junbin Xiao, Angela Yao, Yicong Li, and Tat~Seng Chua.
\newblock Can i trust your answer? visually grounded video question answering.
\newblock In \emph{CVPR}, 2024.

\bibitem[Xu et~al.(2019)Xu, He, Plummer, Sigal, Sclaroff, and Saenko]{xu2019multilevel}
Huijuan Xu, Kun He, Bryan~A Plummer, Leonid Sigal, Stan Sclaroff, and Kate Saenko.
\newblock Multilevel language and vision integration for text-to-clip retrieval.
\newblock In \emph{AAAI}, 2019.

\bibitem[Yang et~al.(2023)Yang, Nagrani, Laptev, Sivic, and Schmid]{yang2024vidchapters}
Antoine Yang, Arsha Nagrani, Ivan Laptev, Josef Sivic, and Cordelia Schmid.
\newblock Vidchapters-7m: Video chapters at scale.
\newblock 2023.

\bibitem[Ye et~al.(2024)Ye, Xu, Xu, Ye, Yan, Zhou, Wang, Hu, Shi, Shi, et~al.]{ye2023mplug}
Qinghao Ye, Haiyang Xu, Guohai Xu, Jiabo Ye, Ming Yan, Yiyang Zhou, Junyang Wang, Anwen Hu, Pengcheng Shi, Yaya Shi, et~al.
\newblock mplug-owl: Modularization empowers large language models with multimodality.
\newblock In \emph{CVPR}, 2024.

\bibitem[Yuan et~al.(2019)Yuan, Mei, and Zhu]{yuan2019find}
Yitian Yuan, Tao Mei, and Wenwu Zhu.
\newblock To find where you talk: Temporal sentence localization in video with attention based location regression.
\newblock In \emph{AAAI}, 2019.

\bibitem[Zhang et~al.(2023)Zhang, Li, and Bing]{zhang2023video}
Hang Zhang, Xin Li, and Lidong Bing.
\newblock Video-llama: An instruction-tuned audio-visual language model for video understanding.
\newblock In \emph{EMNLP}, 2023.

\bibitem[Zhang et~al.(2020{\natexlab{a}})Zhang, Sun, Jing, and Zhou]{zhang2020span}
Hao Zhang, Aixin Sun, Wei Jing, and Joey~Tianyi Zhou.
\newblock Span-based localizing network for natural language video localization.
\newblock In \emph{ACL}, 2020{\natexlab{a}}.

\bibitem[Zhang et~al.(2024)Zhang, Han, Liu, Gao, Zhou, Hu, Yan, Lu, Li, and Qiao]{zhang2023llama}
Renrui Zhang, Jiaming Han, Chris Liu, Peng Gao, Aojun Zhou, Xiangfei Hu, Shilin Yan, Pan Lu, Hongsheng Li, and Yu~Qiao.
\newblock Llama-adapter: Efficient fine-tuning of language models with zero-init attention.
\newblock In \emph{ICLR}, 2024.

\bibitem[Zhang et~al.(2020{\natexlab{b}})Zhang, Peng, Fu, and Luo]{zhang2020learning}
Songyang Zhang, Houwen Peng, Jianlong Fu, and Jiebo Luo.
\newblock Learning 2d temporal adjacent networks for moment localization with natural language.
\newblock In \emph{AAAI}, 2020{\natexlab{b}}.

\bibitem[Zhu et~al.(2024{\natexlab{a}})Zhu, Lin, Ning, Yan, Cui, Wang, Pang, Jiang, Zhang, Li, et~al.]{zhu2023languagebind}
Bin Zhu, Bin Lin, Munan Ning, Yang Yan, Jiaxi Cui, HongFa Wang, Yatian Pang, Wenhao Jiang, Junwu Zhang, Zongwei Li, et~al.
\newblock Languagebind: Extending video-language pretraining to n-modality by language-based semantic alignment.
\newblock In \emph{ICLR}, 2024{\natexlab{a}}.

\bibitem[Zhu et~al.(2024{\natexlab{b}})Zhu, Chen, Shen, Li, and Elhoseiny]{zhu2023minigpt}
Deyao Zhu, Jun Chen, Xiaoqian Shen, Xiang Li, and Mohamed Elhoseiny.
\newblock Minigpt-4: Enhancing vision-language understanding with advanced large language models.
\newblock In \emph{ICLR}, 2024{\natexlab{b}}.

\bibitem[Zhu et~al.(2020)Zhu, Ahuja, Juan, Wei, and Reddy]{zhu2020question}
Ming Zhu, Aman Ahuja, Da-Cheng Juan, Wei Wei, and Chandan~K Reddy.
\newblock Question answering with long multiple-span answers.
\newblock In \emph{EMNLP}, 2020.

\end{thebibliography}
}

% %%%%%%%%%%%%%%%%%%%%%%%%%%%%%%%%%%%%%%%%%%%%%%%%%%%%%%%%%%%%

\appendix

\newpage

\appendix

\begingroup
\hypersetup{colorlinks=false, linkcolor=black}
\hypersetup{pdfborder={0 0 0}}
\part{} % Start the appendix part
\parttoc % Insert the appendix TOC
\endgroup

\section{Changelog}
\label{changelog}

\subsection{Version 2}
\begin{itemize}
    \item We have fixed some bugs in the evaluation code, resulting in slight differences compared to the previous release. The issue was that 149 samples were not evaluated in the previous version, and these have now been included in the new update.
\end{itemize}

\section{Additional documentation and resources}
\subsection{Limitations}
\label{limit}
Despite these advancements, our dataset does exhibit certain limitations, largely stemming from inherited biases from the source datasets:
\begin{itemize}
    \item Currently, we only address scenarios where both the question and the answer span a single time duration. Given a question, the annotated time span must be a single, continuous duration, which might be limiting for all scenes.
    \item The presence of noisy or inaccurate annotations in the source datasets, including captions and timestamps, poses a challenge. Despite our efforts, some of these errors could not be automatically filtered out. The extent of this issue is detailed in the qualitative visualization conducted by our human reviewers, as presented in supplementary.
    \item The average duration of ground truth events in our dataset is relatively long. This characteristic has the unintended consequence of hindering the models' ability to detect and analyze fine-grained actions within shorter video segments.
\end{itemize}

These drawbacks highlight areas for potential improvement and indicate the necessity for ongoing refinement to ensure the creation of more accurate and unbiased video language models.

\subsection{Social Impact}
\label{social_impact}
Though we provide an assessment of temporal reasoning and moment localization, the types and scene diversity are still limited. We inherit the video classes from the two source video datasets, which may not be sufficient for a comprehensive assessment of all kinds of temporal reasoning. This limitation could introduce a bias. 

For both curated data and video data, they do not contain any personally identifiable information. Besides, some of the video samples in the source datasets might be slightly uncomfortable depending on the viewer. For example, some videos discuss tattoos and piercings, and some of them present news about social events including demonstrations or war reports. However, we only release the data of curated question-answer and time span. We are not responsible for the release and maintenance of video data.

\subsection{Data source links}
\label{links}
Author's email: \href{r12942106@ntu.edu.tw}{r12942106@ntu.edu.tw} \\
Project page: \url{https://rextime.github.io/} \\
Huggingface dataset: \url{https://huggingface.co/datasets/ReXTime/ReXTime} \\
Github (code, data): \url{https://github.com/ReXTime/ReXTime} \\
Croissant: \url{https://huggingface.co/api/datasets/ReXTime/ReXTime/croissant}
\subsection{License}
\label{license}
Our generated data is released under \href{https://creativecommons.org/licenses/by-nc-sa/4.0/}{CC BY-NC-SA 4.0} license. Our code provided in Github is released under \href{https://opensource.org/license/mit}{MIT} license.

\subsection{Author statement}
\label{statement}
As the author of this work, we take full responsibility for any rights violations, including intellectual property rights. We confirm that all data used complies with applicable licenses and legal requirements, and all external sources have been properly credited and permissions obtained. This statement acknowledges our accountability and adherence to relevant data and copyright regulations.

\subsection{Maintenance plan}
\label{maintenance}
We will host and continuously update our data through various release sources, including GitHub (code and data), Huggingface (datasets), our project page, and the Eval.AI challenge.

\subsection{Digital object identifier~(DOI)}
\label{DOI}
You can find the digital object identifier in our citation block on Huggingface dataset page: \\
\url{https://huggingface.co/datasets/ReXTime/ReXTime}

\subsection{Annotation instruction}
\label{anno_instruction}
We provide the link to the slide which is used in the annotation process as an instruction. Note that the used language in the slide is Chinese. Slide: \url{https://docs.google.com/presentation/d/1-wgWYaWF-ZIa1YBSyPGc5p5TqTGXorxqkYZhOGIqqBg/edit?usp=sharing}

\section{Additional implementation details}
\subsection{Source Datasets}
\label{subsec:source_dataset}

\paragraph{ActivityNet}
ActivityNet is a comprehensive large-scale video benchmark designed to advance the field of human activity recognition by addressing the limitations of current computer vision algorithms. ActivityNet offers a diverse collection of complex human activities that reflect everyday life. The dataset encompasses 203 distinct activity classes, each with an average of $137$ untrimmed videos, and features approximately $1.41$ activity instances per video. This results in a substantial total of 849 video hours. ActivityNet supports various evaluation scenarios, including untrimmed video classification, trimmed activity classification, activity detection and dense video captions, making it a valuable resource for comparing and improving algorithms for human activity understanding.

\paragraph{QVHighlights}
QVHighlights dataset addresses the challenge of detecting video moments and highlights based on natural language (NL) queries, an underexplored area due to a lack of annotated data. It includes over $10,000$ YouTube videos on various topics, each annotated with NL queries, relevant moments, and five-point saliency scores. This enables the development and evaluation of systems for detecting relevant moments and highlights. QVHighlights focused on user-created lifestyle vlog videos on YouTube. These videos, made by users worldwide, showcase various events and aspects of their lives, including everyday activities and travel. Captured with different devices (\ie, smartphones, GoPro) and view angles (\ie, first-person, third-person), they present significant challenges to computer vision systems. To enhance dataset diversity, we also included news videos with substantial ``raw footage'', covering serious topics like natural disasters and protests. We used queries such as ``daily vlog'', ``travel vlog'', and ``news hurricane'' to harvest videos from YouTube, selecting top results between 5-30 minutes long, uploaded after $2016$ for better visual quality, and filtering out videos with low view counts or high dislike ratios. These raw videos were then segmented into 150-second clips for annotation.

\subsection{Filter}
\label{subsec:filter}
The initial stage involves filtering out samples unsuitable for conversion into a temporal reasoning format. A temporal reasoning conversation sample requires a complex scene with sequential events occurring in it. Also, we need information which describes segments in detail instead of an overall summary of a whole video. Last but not the least, we want the sample source to contain as much information as possible. That's why we need a filter to select a proper sample source.
For data originating from QVHighlights, we eliminate samples wherein the video content represents a single, continuous event. Specifically, this refers to videos where the answer span encompasses the entire duration, from start to finish. Also, we exclude samples if a query happens several times in the video, which indicates a routine and repeated behavior. In contrast, we apply a distinct set of criteria for filtering for samples from ActivityNet. First, samples with an event duration exceeding 80\% of the total video length are discarded. This criterion helps ensure a diverse range of events within each video. Second, samples where the cumulative duration of all segments is less than 60\% of the video's total length are regarded as insufficiently detailed (``sparse captioning'') and are therefore excluded. This is due to potential information deficits in such samples. Third, we perform a clustering of event intervals, applying a threshold of 10 seconds. Intervals separated by gaps exceeding this threshold are considered discontinuous and are segmented into distinct groups. From these groups, we select the one with the highest event count for the generation of question-answer pairs, ensuring richness in temporal reasoning content.

\subsection{Cost estimation}
\paragraph{Test data generation and Verification}
\label{subsec:verify_cost}
We take $1000$ samples as an example. One person can review 60 samples per hour. Generating $1000$ samples with GPT-4 costs about $35\$$. At a minimum hourly rate of $6\$$, the total cost for $1000$ samples, including human verification, is about $135\$$. Conversely, creating 20 natural language question-answer pairs for video content takes about one hour. Thus, generating $1000$ samples would require $50$ hours, costing $300\$$ in total. Our pipeline can create video QA data much more efficiently, at only \textbf{$45\%$} of the total cost.

\paragraph{Training data generation}
\label{subsec:training_cost}
We take $1000$ samples as an example. Generating $1000$ samples with GPT-4 costs about $35\$$. The total cost for generating $1000$ training samples is about $35\$$. Conversely, creating 20 natural language question-answer pairs for video content takes about one hour. Thus, generating $1000$ samples would require $50$ hours, costing $300\$$ in total. Our pipeline can create reasoning-across-time training video QA data much more efficiently, at a bit more than \textbf{$10\%$} of the total cost.

\subsection{Computing resources}
\label{subsec:compute}
All of our fine-tuning experiments are done with an Nvidia RTX-3090 24G GPU.

\subsection{Training details and hyper-parameters}
\label{subsec:training_details}
We report the training details and hyper-parameters in this section. Overall, we will follow the setting provided by the original papers or official Github setting. However, to fine-tune grounding video-language models such as~\citep{huang2023vtimellm, ren2023timechat} on resource as reported in~\cref{subsec:compute}, we will apply LoRA~\citep{hu2022lora} fine-tuning and reduce batch size.

\paragraph{UniVTG~\citep{lin2023univtg}}
We follow the single-gpu training script~\footnote{\url{https://github.com/showlab/UniVTG/blob/main/scripts/qvhl_pretrain.sh}} provided by UniVTG official implementation with learning rate 1e-4, clip lengths 2, batch size 32, epochs 200 and hidden dimension 1024. We load the weight pre-trained on several datasets released by UniVTG official implementation for both zero-shot moment retrieval and fine-tuning experiments.

\paragraph{CG-DETR~\citep{moon2023correlation}}
We load the weight pre-trained on QVHighlights released by CG-DETR official implementation for zero-shot moment retrieval. We follow the single-gpu training script~\footnote{\url{https://github.com/wjun0830/CGDETR/blob/main/cg_detr/scripts/train.sh}} provided by CG-DETR official implementation to train on our generated data.

\paragraph{VTimeLLM~\citep{huang2023vtimellm}}
To evaluate zero-shot performance, we load the stage 3 model weight from the VTimeLLM official implementation. We assess moment retrieval and VQA (Visual Question Answering) performance separately. For moment retrieval, we prompt the model with ``Can you pinpoint when and...'' followed by the question sentence, and extract the time token from the predicted sentence. For zero-shot VQA evaluations, we concatenate four options after the prefix ``From <ss> to <ee>, <option>'' as four predictions, here <ss> and <ee> is ground truth span. Then we calculate the sequence probability for each, and select the maximum probability as the VQA prediction.

For fine-tuning experiments, we follow the tuning strategy provided by VTimeLLM. Starting with the stage 3 model weight, we add a new LoRA adapter, tune on our generated training dataset, and merge the adapter during inference. We use the hyper-parameters from the original paper: a learning rate of 1e-4, number of video frames of 100, LoRA rank of 64, LoRA alpha of 128, training for 2 epochs, with a batch size of 8 and gradient accumulation steps of 16. For fine-tuned evaluation, we first predict the whole sentence given a question sentence and extract the predicted time tokens <ss> and <ee>. We then concatenate the four options after the predicted answer span ``From <ss> to <ee>, <option>'' as four predictions, calculate the sequence probability, and choose the maximum one for VQA and GQA (Grounding VQA) prediction. Here we provide a python pesudo as a demonstration:

\noindent\rule{\textwidth}{1pt}
\textbf{Pseudo code: }
\label{pseudo_code}
This is a python pseudo code for the assessment of grounding multi-choice VQA.
\begin{lstlisting}[language=Python]
def extract_time_token(string):
    # string: From ss to ee, the girl is ....
    pattern = r"\s+(\d+)\s+to\s+(\d+)"
    matches = re.findall(pattern, string)
    return matches

def get_predicted_score(logits, labels):
    # Get label start index and end index
    start_idx, end_idx = ...
    scores = nn.CrossEntropyLoss(logits[start_idx:end_idx+1], \\
                                labels[start_idx:end_idx+1])
    return scores

def concat(question, predicted_time_tokens, option)
    # question + 'From ss to ee' + option.
    return question + predicted_time_tokens + option

# Input: question(string), options(string in list, lenght==4)

# Time tokens prediction (Moment localization)
output = model.generate(question)
# Decode to natural language
response = tokenizer.decode(outputs)
# Extract time tokens
predicted_time_tokens = extract_time_token(response)

# Concatenate predicted_time_tokens with each option.
# From ss to ee, <Option>.
inputs = []
for i in range(4):
    inputs.append(concat(question, predicted_time_tokens, options[i]))
inputs = tokenizer.encode(inputs)

# Multi-choice prediction (VQA)
# input_ids.shape==(4, batch_max_lenghts) for 4 options
output = model(**inputs['input_ids'])
# Compute the mean of labels sequence log-probability.
scores = get_predicted_score(output['logits'], inputs['labels'])
# Find the one with largest crossentropyloss as predicted answer
predicted_answer = transition_scores.max()
\end{lstlisting}

\paragraph{TimeChat~\citep{ren2023timechat}}
For the zero shot setting, we evaluate the checkpoints from the TimeChat official implementation. We also assess moment retrieval and VQA (Visual Question Answering) separately. For the first task we follow their prompt for temporal retrieval, and parse model's response to obtain the timestamps prediction. The evaluation process for zero-shot VQA for TimeChat is the same as that for VTimeLLM.

When fine-tuning TimeChat on our proposed dataset, we start from fine-tuned checkpoints provided by TimeChat and follow their instruction fine-tuning settings. Specifically, we use LoRA with a rank of 32, alpha of 128. We train the model with a learning rate of 3e-5, batch size of 8, and gradient accumulation steps of 8 for 3 epochs. The number of frames used in each video is 96. To evaluate the performance after fine-tuning, we use the same evaluation protocol as that we used for VTimeLLM.

\subsection{Counting temporal reasoning QAs}
\label{subsec:count_temp}
We compare \ours with Ego4D-NLQ and NExTGQA. For metrics like \textbf{average certificate lengths (C.L.)} and \textbf{question-answer intersection of union (QA-IoU)}, we follow the methodology from~\citet{mangalam2024egoschema}, manually annotating at least two hours of human effort for each dataset. A screenshot of the labeling GUI tool is provided in \cref{subsec:gui}.

To determine the \textbf{number of reasoning across time samples}, we count the total queries with ``before/after'' in Ego4D-NLQ and the samples of the ``temporal'' type in NExTGQA. We exclude other cases where the question time span completely overlaps with the answer time span, as they do not qualify as "reasoning across time."

\subsection{GUI}
\label{subsec:gui}
\begin{figure}[!t]
  \centering
  \includegraphics[width=0.95\textwidth]{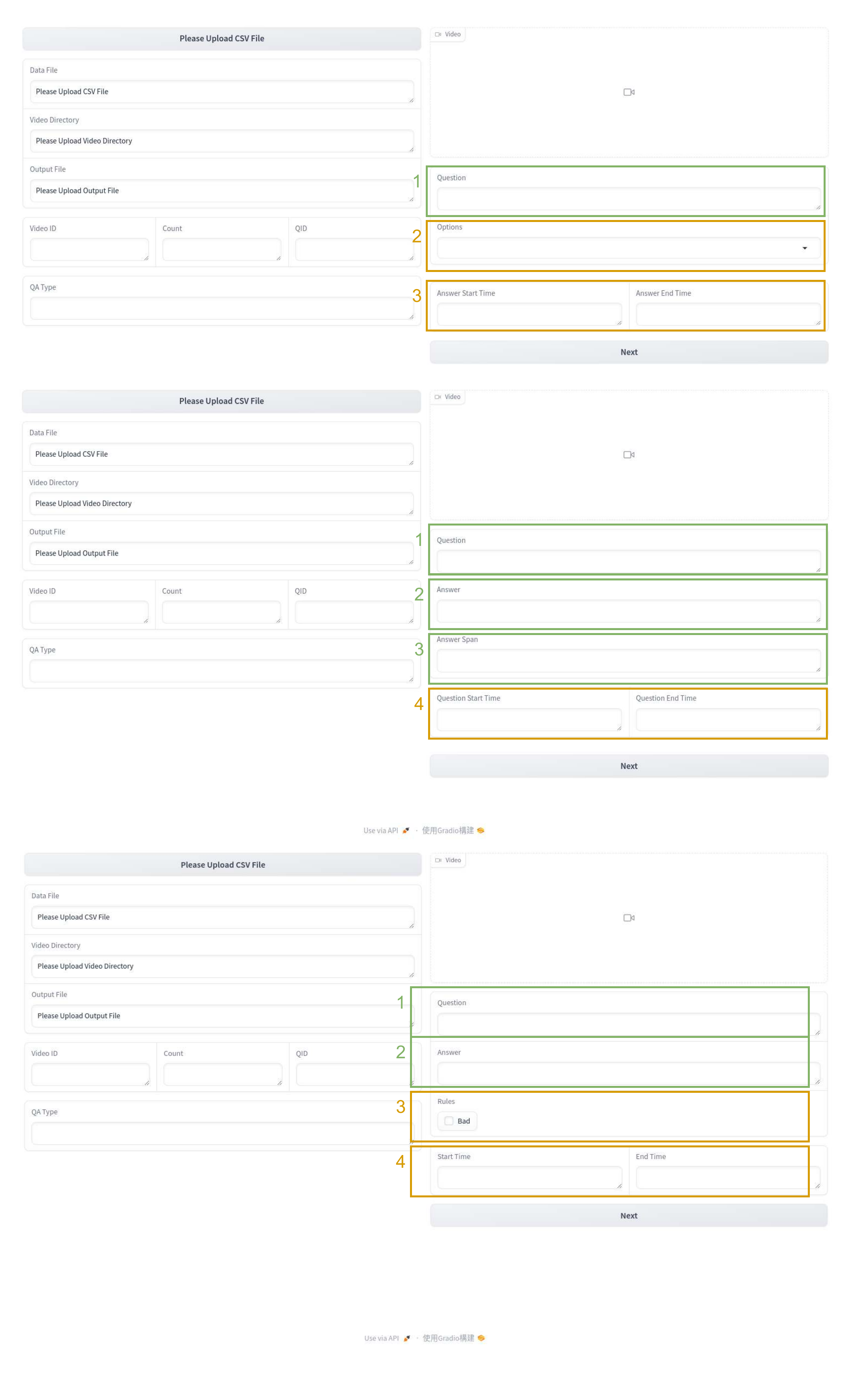}
  \caption{We show the GUIs for different annotation / verification processes.}
  \label{fig:guis}
\end{figure}
To facilitate efficient annotation, we have developed a Gradio graphical user interface (GUI).\footnote{\url{https://www.gradio.app/}} Here we provide three types of annotation tool, human time span annotation and verification, human question span annotation and human performance annotation. Here we show in~\cref{fig:guis}. The first one is for human time span annotation. The annotators are responsible for assessing each question-answer pair to ensure logical coherence and alignment to the video content. Additionally, they need to provide the time span of the answer, which will be used as ground truth in the following. The second one is for human question span annotation, given question, answer and answer span, the annotators need to find a span which is relevant to the question event. This is for the assessment of average certificate lengths (C.L.) and question-answer intersection of union (QA-IoU). The third one is for human performance experiment, given question and four options, the participants need to find not only the answer from the four options but also an answer span which is relevant to the selected answer.The green area indicates what an annotator will get, and the orange area indicates what an annotator need to answer.

\subsection{Prompts}
\subsubsection{ActivityNet event generation}
\begin{lstlisting}
Following the steps below to evaluate the causality between two events in the video.
First, find two events from different timestamps which have strong causality.
Second, evaluate the causality between the two events according to the following criteria:
a. Directness: How directly does one event lead to the next?
b. Necessity: Is the subsequent event a necessary consequence of the previous one?
c. Intentionality: Determine if the first event is deliberately executed to cause the second event.
d. Purpose: Assess whether the first event is conducted with the primary goal of leading to the second event.
Scoring Method for criteria a to d (Score 0-3 for each criterion):
0: Weak causal relationship.
1: Moderate causal relationship.
2: Strong causal relationship.
3. Definite causal relationship.

e. Similarity: Assess whether the two events are just repeated actions or not.
Scoring Method for criterion e (Score 0-3):
0: Totally different action.
1: Slightly same action with event progression.
2: Partially same action with little event progression.
3: Totally same action without event progression.

Sequential video captioning:
<CAPTIONS>
Provide a brief and concise explanation of the score you give to each criterion.
<Provide your explanation here>
Finish the result json according to your evaluation:
```json{
    "event1": "<EVENT1>",
    "event1_timestamp": [start, end],
    "event2": "<EVENT2>",
    "event2_timestamp": [start, end],
    "Directness": <DIRECTNESS>,
    "Necessity": <NECESSITY>,
    "Intentionality": <INTENTIONALITY>,
    "Purpose": <PURPOSE>,
    "Similarity": <SIMILARITY>
}```
\end{lstlisting}

\subsubsection{QVHighlights event generation}
\begin{lstlisting}
These are frames from a video.
Find out a behavior in the video which is caused by the pivotal event "<QUERY>" or a behavior which leads to the pivotal event."
If there isn't any behavior that is caused by or leads to the pivotal event, return "none".
Your response should be in json format as the following.
{
    "explain": <A brief explanation according to the video and instruction>,
    "cause": <The behavior leads to the pivotal event>,
    "cause-relevant": <Does the cause have strong temporal-causality with the pivotal event? yes or no.>,
    "cause-alignment": <How well the cause is aligned with the video? high, medium, low>,
    "effect": <The behavior caused by the pivotal event>,
    "effect-relevant": <Does the effect have strong temporal-causality with the pivotal event? yes or no.>,
    "effect-alignment": <How well the effect is aligned with the video? high, medium, low>
}
\end{lstlisting}

\subsubsection{Sequential QA generation}
\begin{lstlisting}
Sequential video captioning:
<CAPTIONS>

Find two continuous events in the video captions from different timestamps.
Construct a temporal related question and answer based on the two events.
Examples:
(Pre-event) Jack wakes up. (Post-event) Jack brushes his teeth.
Type1. Question (pre-event): What does Jack do after waking up? Answer (post-event): Jack brushes his teeth.
Type2. Question (post-event): What does Jack do before brushing his teeth? Answer (pre-event): Jack wakes up.

Provide a brief and concise explanation.
<Your brief explanation here>
Finish the result json according to your explanation:
```json{
    "pre-event": "<EVENT1>",
    "pre-event_timestamp": [start, end],
    "post-event": "<EVENT2>",
    "post-event_timestamp": [start, end],
    "Type1": {
        "Question": "<QUESTION>",
        "Answer": "<ANSWER>"
    },
    "Type2": {
        "Question": "<QUESTION>",
        "Answer": "<ANSWER>"
    }
}```
\end{lstlisting}

\subsubsection{Cause-effect QA generation}
\begin{lstlisting}
This is a cause-effect relationship. The event "<EVENT1>" causes the event "<EVENT2>".
Please construct 2 types of questions and answers based on the cause-effect relationship.
Examples:
(Cause) A girl falls off a bike. (Effect) She is injured.
Type1. Question (cause): What does the girl falling off the bike lead to? Answer (effect): She is injured.
Type2. Question (effect): Why is the girl injured? Answer (cause): She falls off the bike.

Provide a brief and concise explanation.
<Your brief explanation here>
Finish the result json according to your explanation:
```json{
    "Type1": {
        "Question": "<QUESTION>",
        "Answer": "<ANSWER>"
    },
    "Type2": {
        "Question": "<QUESTION>",
        "Answer": "<ANSWER>"
    }
}```
\end{lstlisting}

\subsubsection{Means-to-an-end QA generation}
\begin{lstlisting}
This is a means-to-an-end relationship. The event "<EVENT1>" is a means to achieve the event "<EVENT2>".
Please construct a question and an answer based on the means-to-an-end relationship.
Examples:
(Means) Mixing flour and water. (End) Make dough.
Type1. Question (end): How do we make dough? Answer (means): By mixing flour and water.
Type2. Question (means): Why do we mix flour and water? Answer (end): To make dough.

Provide a brief and concise explanation.
<Your brief explanation here>
Finish the result json according to your explanation:
```json{
    "Type1": {
        "Question": "<QUESTION>",
        "Answer": "<ANSWER>"
    },
    "Type2": {
        "Question": "<QUESTION>",
        "Answer": "<ANSWER>"
    }
}```
\end{lstlisting}

\subsubsection{QA verification}
\begin{lstlisting}
You are tasked with verifying if the question-answer pair is logically correct.
Provide your explanation of the verification result.
Provide a score from 0 to 3 to indicate the correctness of the question-answer pair.
Scoring Method:
0: Incorrect question-answer pair.
1: Relevant but not logically correct.
2: Partially correct.
3: Completely correct.

Case 1:
Question: <QUESTION1>
Answer: <ANSWER1>

Case 2:
Question: <QUESTION2>
Answer: <ANSWER2>

Provide a brief and concise explanation.
<Your brief explanation here>
Finish the result json according to your explanation:
```json{
    "case_1_score": <SCORE>,
    "case_2_score": <SCORE>
}```
\end{lstlisting}

\subsubsection{Options generation}
\begin{lstlisting}
You are tasked with generating high-quality, incorrect options for a given question-answer pair.
The options should be logically consistent and correct, but they must be different from the correct answer.
Note that the generated options should not be a interpretations or variations of the correct answer in any way.

Here are the question-answer pair:
Question: <QUESTION>
Answer: <ANSWER>

Provide a brief and concise explanation.
<Your brief explanation here>
Finish the result json according to your explanation:
```json{
    "correct_answer": "<ANSWER>",
    "wrong_answer_1": "<OPTION1>",
    "wrong_answer_2": "<OPTION2>",
    "wrong_answer_3": "<OPTION3>"
}```
\end{lstlisting}
\section{Additional experiment results}
\subsection{Qualitative Results}
\label{subsec:quality_result}
\begin{figure}[!t]
    \centering
    \begin{minipage}{\textwidth}
        \centering
        \includegraphics[width=0.975\textwidth]{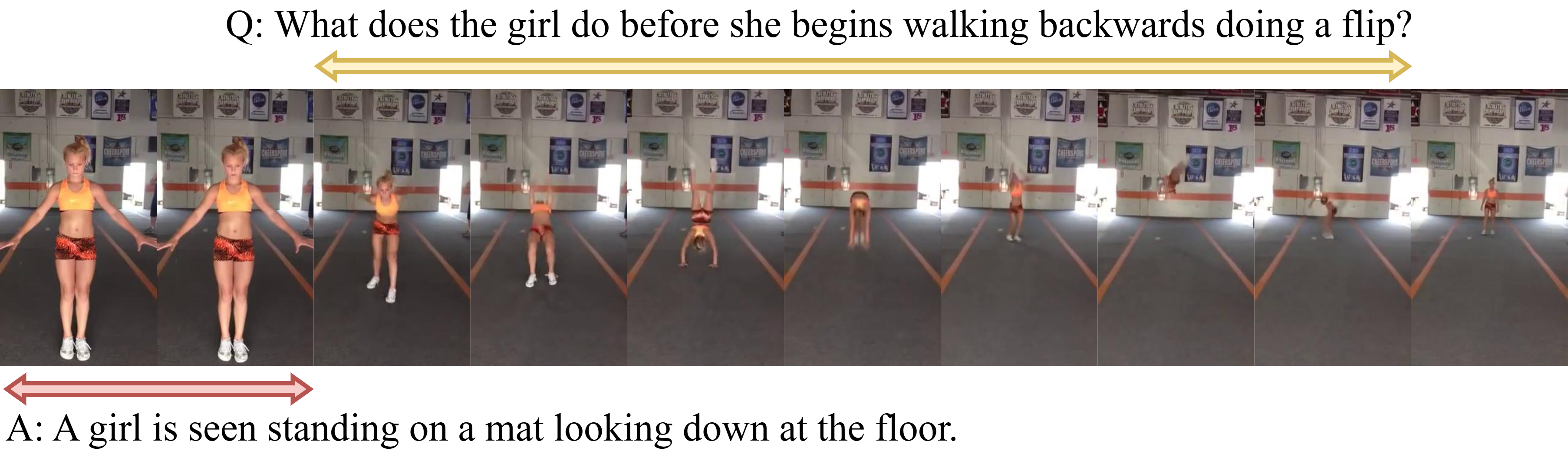}
        \label{fig:q1}
    \end{minipage}
    \hfill
    \begin{minipage}{\textwidth}
        \centering
        \includegraphics[width=0.975\textwidth]{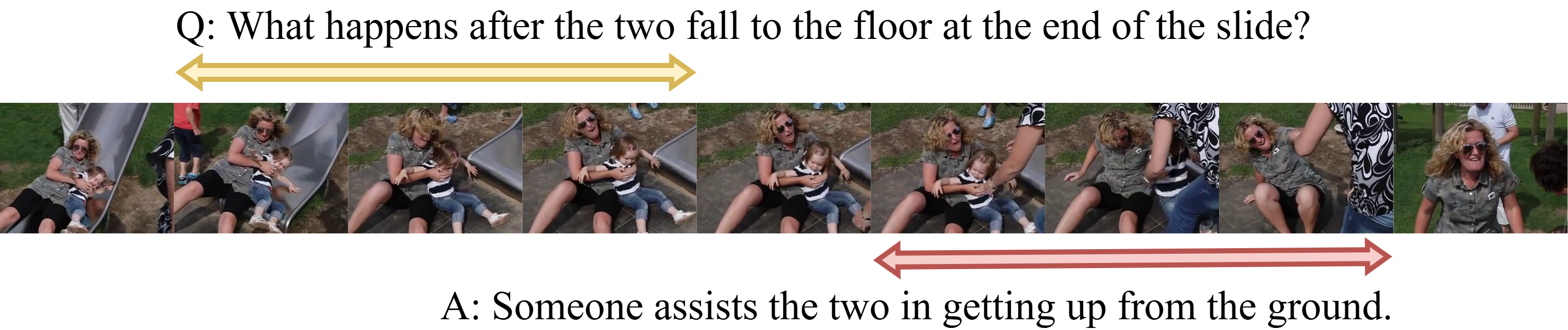}
        \label{fig:q2}
    \end{minipage}
    \hfill
    \begin{minipage}{\textwidth}
        \centering
        \includegraphics[width=0.975\textwidth]{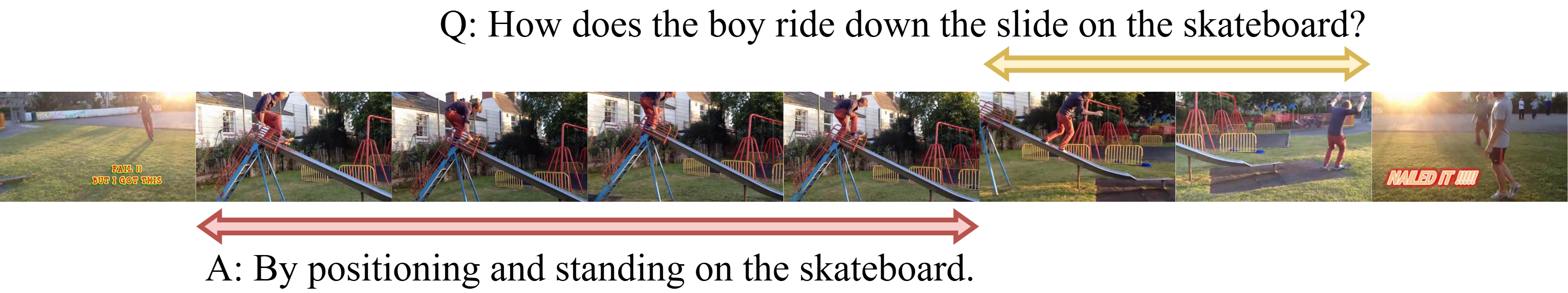}
        \label{fig:q3}
    \end{minipage}
    \hfill
    \begin{minipage}{\textwidth}
        \centering
        \includegraphics[width=0.975\textwidth]{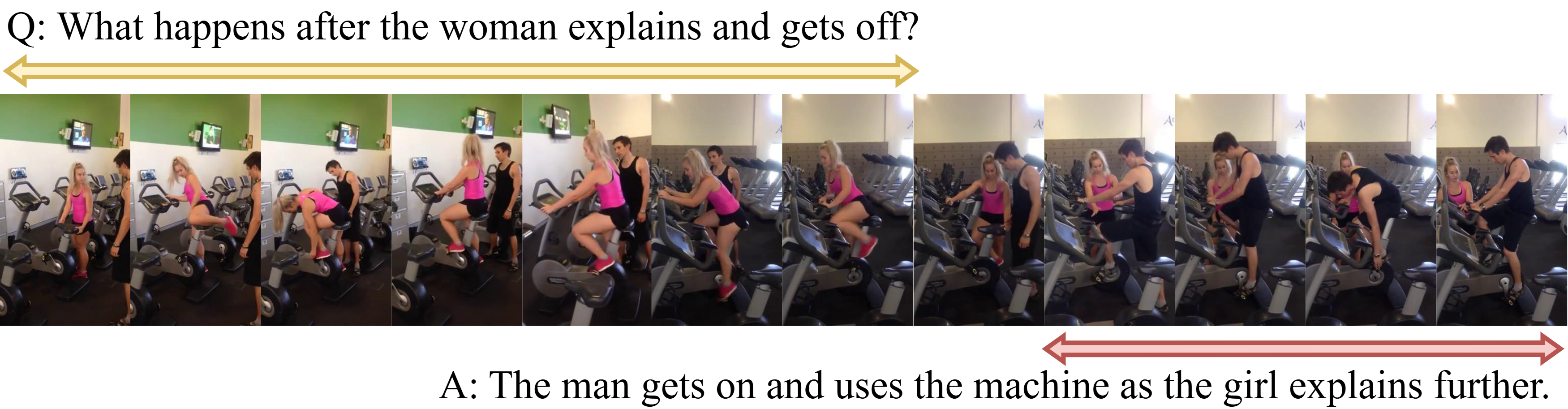}
        \label{fig:q4}
    \end{minipage}
    \hfill
    \begin{minipage}{\textwidth}
        \centering
        \includegraphics[width=0.975\textwidth]{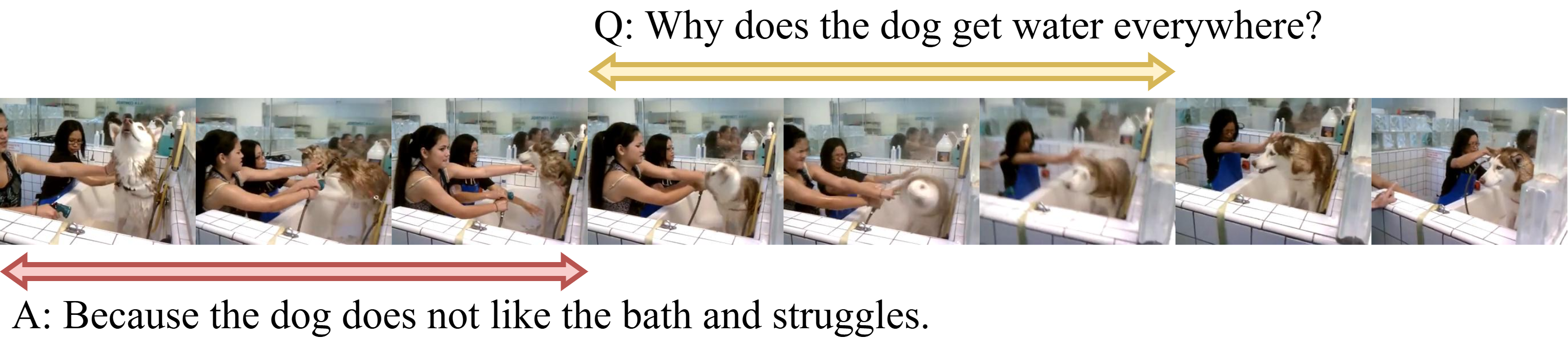}
        \label{fig:q5}
    \end{minipage}
    \hfill
    \begin{minipage}{\textwidth}
        \centering
        \includegraphics[width=0.975\textwidth]{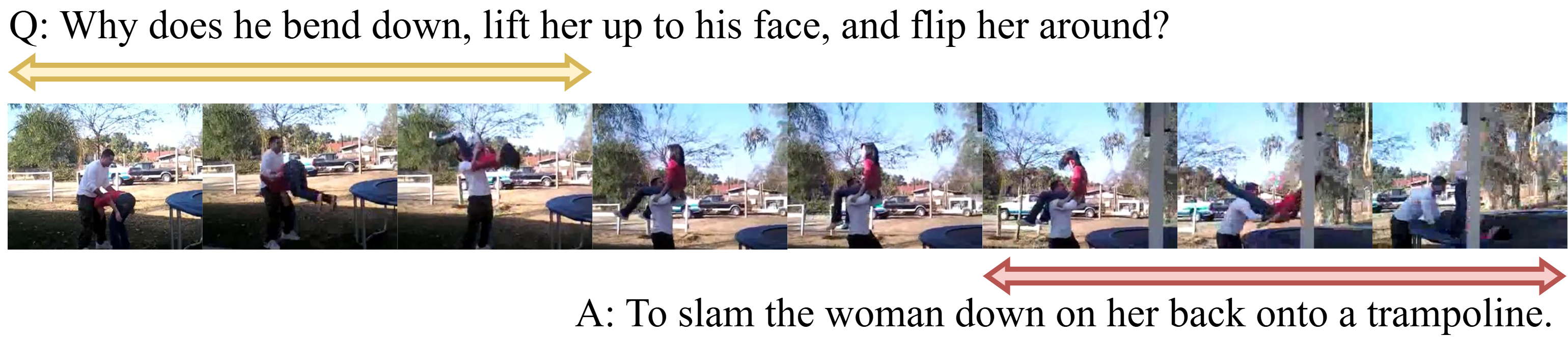}
        \label{fig:q6}
    \end{minipage}
    \caption{We show the qualitative results of generated data.}
    \label{fig:qualitative}
\end{figure}
We present six generated data examples in~\cref{fig:qualitative}, categorized into three types: \textit{sequential}, \textit{cause-effect}, and \textit{means-to-an-end}. For each category, we include one example where the question event happens first and another where the answer event happens first.

\subsection{Teaser examples}
\label{subsec:teaser_cases}

Here we provide the inference case depicted in teaser on GPT-4V~\citet{achiam2023gpt}, Gemini-1.5-Pro~\citet{team2023gemini}, Claude3-Opus~\citet{Clude3} and Reka-Core~\citet{Reka}.

\subsubsection{GPT-4V}
\textbf{Number of Frames:} 50 \\
\textbf{Prompt Input:}
\begin{lstlisting}
According to the video, answer the following questions:
How can we cut the tomato more efficiently?
\end{lstlisting}

\textbf{Response:}
\begin{lstlisting}
To cut the tomato more efficiently, the video demonstrates a technique where you can use a plate as a guide to slice the tomato. Here are the steps based on the images provided:

    1. Place the tomato on a chopping board.
    2. Cover the tomato with a flat, slightly deep plate, upside down.
    3. Hold the plate firmly in place over the tomato.
    4. Using a sharp knife, slice horizontally through the gap between the chopping board and the bottom of the plate. This allows you to slice the tomato into even sections while keeping your fingers safely away from the knife.
    5. Carefully lift the plate to reveal the evenly sliced tomato segments beneath.

This method is particularly useful for slicing multiple tomatoes quickly and uniformly, and it also helps in keeping the slices contained, making it cleaner and more organized.
\end{lstlisting}

\subsubsection{Gemini-1.5-Pro}
\textbf{Number of Frames:} 50 \\
\textbf{Prompt Input:}
\begin{lstlisting}
According to the video, please describe the video content.
\end{lstlisting}
\textbf{Response:}
\begin{lstlisting}
A man is in a kitchen. He is preparing a meal. He is cutting tomatoes and there are other ingredients on the counter, such as meat, green beans, and oil. The man is wearing a hat and a light blue shirt.
\end{lstlisting}
\textbf{Prompt Input:}
\begin{lstlisting}
According to the video, answer the following questions: How can we cut the tomato more efficiently?
\end{lstlisting}
\textbf{Response:} \\
\begin{lstlisting}
The video does not provide any information on how to cut tomatoes more efficiently. The video only shows a person cutting tomatoes.
\end{lstlisting}

\subsubsection{Claude3-Opus}
\textbf{Number of Frames:} 20 \\
\textbf{Prompt Input:}
\begin{lstlisting}
According to the video, answer the following questions: How can we cut the tomato more efficiently?
\end{lstlisting}
\textbf{Response:}
\begin{lstlisting}
The video demonstrates using a plate or flat surface to help guide the knife and cut the tomato into even slices more efficiently. By holding the tomato against the plate while slicing downward with the knife, it allows for making consistent, thin slices quickly without the tomato rolling around on the cutting board. The plate acts as a stable guide to help control the slicing motion and produce uniform tomato slices with less effort compared to trying to slice the tomato freehand.
\end{lstlisting}

\subsubsection{Reka-Core}
\textbf{Number of Frames:} Unknown \\
\textbf{Prompt Input:}
\begin{lstlisting}
According to the video, answer the following questions: How can we cut the tomato more efficiently? 
\end{lstlisting}
\textbf{Response:}
\begin{lstlisting}
The video demonstrates a method for slicing a tomato. To cut the tomato more efficiently, you should start by slicing off the bottom of the tomato to create a stable base. Then, slice the tomato horizontally into even rounds, depending on your preference. If you want to dice the tomato, after slicing it into rounds, stack the slices and cut across them to create cubes. It is important to use a sharp knife for clean cuts and to hold the tomato firmly with a claw grip to avoid cutting your fingers.
\end{lstlisting}
\textbf{Prompt Input:}
\begin{lstlisting}
According to the video, answer the following questions: Why do we hold up a plate?
\end{lstlisting}
\textbf{Response:}
\begin{lstlisting}
In the video, the act of holding up a plate can have several purposes. It could be to inspect the plate for cleanliness, to show the plate to someone else, or to demonstrate a technique or feature of the plate, such as a logo or a special design on the bottom. It could also be part of a cooking demonstration or instructional video where the person is explaining the next step or ingredient to the audience.
\end{lstlisting}

\subsubsection{GPT-4o}
\textbf{Number of Frames:} 50 \\
\textbf{Prompt Input:}
\begin{lstlisting}
According to the video, answer the following questions: How can we cut the tomato more efficiently?
\end{lstlisting}
\textbf{Response:}
\begin{lstlisting}
Based on the sequence of images from the video, here is how you can cut a tomato more efficiently:
**Use a Sharp Knife: Ensure your knife is sharp. The video shows the person demonstrating the sharpness of the knife using the edge of a plate.
**Sharpen the Knife**: If your knife is not sharp enough, you can use the unglazed bottom rim of a ceramic plate to sharpen it. Run the knife's edge along the rim at an angle to hone the blade. This can be a handy trick if you don't have a knife sharpener.
**Proper Technique**:
    1. Start by halving the tomato.
    2. Place the tomato halves on the cutting board with the flat side down for stability.
    3. Slice the tomato into even pieces, keeping your fingers tucked safely away from the blade.
By ensuring your knife is sharp and using effective slicing techniques, you can cut the tomato more efficiently and safely.
\end{lstlisting}

\newpage
\subsection{Open source performance on mini test set}
We show the performance results of open source models on the mini test set. Please refer to~\cref{tab:sup_zeroshot} and~\cref{tab:sup_finetuned}.
\label{subsec:mini_test}
\begin{table}[t]
\caption{Zero-shot performance on mini test set. We report open source models' zero-shot performance on the mini test set compared to the full set.
}
\centering
\newcolumntype{C}{>{\centering\arraybackslash}X}
\newcolumntype{L}{>{\raggedright\arraybackslash}X}
\newcolumntype{R}{>{\raggedleft\arraybackslash}X}
\begin{tabularx}{\columnwidth}{lCCCcc}
\toprule
\multirow{3}{*}{Models} & \multicolumn{3}{c}{Moment Localization} &  \multicolumn{2}{c}{VQA} \\
\cmidrule(lr){2-4}
\cmidrule(lr){5-6}
       & \multirow{2}{*}{mIoU} & R@1  & R@1  & Accuracy(\%) & Accuracy(\%)\\
       && \footnotesize (IoU=0.3) & \footnotesize (IoU=0.5) &  & \footnotesize @ IoU $\geq 0.5$\\
\midrule
UniVTG (zs)  & 28.17 & 41.34 & 26.88 & $-$ & $-$ \\
UniVTG (zs,mini) & 30.18 & 42.00 & 29.33 & $-$ & $-$ \\
CG-DETR (zs) & 23.87 & 31.31 & 19.60 & $-$ & $-$\\
CG-DETR (zs,mini) & 22.53 & 30.00 & 16.67 & $-$ & $-$ \\

\cmidrule(lr){1-6}
VTimeLLM (zs) & 20.14 & 28.84 & 17.41 & 36.16 & $-$\\
VTimeLLM (zs,mini) & 19.37 & 27.67 & 16.00 & 37.33 & $-$\\

TimeChat (zs) & 11.65 & 14.42 & 7.61 & 40.04 & $-$\\
TimeChat (zs,mini) & 13.01 & 16.33 & 7.00 & 38.33 & $-$\\

LITA (zs) & 21.49 & 29.49 & 16.29 & 34.44 & $-$ \\
LITA (zs,mini)& 24.76 & 34.33 & 20.00 & 35.00 & $-$ \\

\bottomrule
\end{tabularx}
\label{tab:sup_zeroshot}
\end{table}
\begin{table}[t]
\caption{Fine-tuned performance on mini test set. We report open source models' fine-tuned performance on the mini test set compared to the full set. }
\centering
\newcolumntype{C}{>{\centering\arraybackslash}X}
\newcolumntype{L}{>{\raggedright\arraybackslash}X}
\newcolumntype{R}{>{\raggedleft\arraybackslash}X}
\begin{tabularx}{\columnwidth}{lCCCcc}
\toprule
\multirow{3}{*}{Models} & \multicolumn{3}{c}{Moment Localization} &  \multicolumn{2}{c}{VQA} \\
\cmidrule(lr){2-4}
\cmidrule(lr){5-6}
       & \multirow{2}{*}{mIoU} & R@1  & R@1  & Accuracy(\%) & Accuracy(\%)\\
       && \footnotesize (IoU=0.3) & \footnotesize (IoU=0.5) &  & \footnotesize @ IoU $\geq 0.5$\\
\midrule
UniVTG (ft)   & 34.63 & 53.48 & 34.53 & $-$ & $-$\\
UniVTG (ft,mini) & 34.82 & 53.00 & 35.33 & $-$ & $-$ \\
CG-DETR (ft) & 26.53 & 39.71 & 22.73 & $-$ & $-$\\
CG-DETR (ft,mini) & 24.98 & 38.00 & 20.33 & $-$ & $-$ \\

\cmidrule(lr){1-6}
VTimeLLM (ft) & 29.92 & 43.96 & 26.13 & 57.58 & 17.13\\
VTimeLLM (ft,mini) & 29.53 & 43.67 & 25.00 & 54.67 & 15.00\\

TimeChat (ft) & 26.29 & 40.13 & 21.42 & 49.46 & 10.92\\
TimeChat (ft,mini) & 27.54 & 38.00 & 21.67 & 52.00 & 11.33\\

\bottomrule
\end{tabularx}
\label{tab:sup_finetuned}
\end{table}

%%%%%%%%%%%%%%%%%%%%%%%%%%%%%%%%%%%%%%%%%%%%%%%%%%%%%%%%%%%%

\end{document}